%% file: FL_Healthcare_ACM.tex
\documentclass[acmlarge]{acmart}

\usepackage[ansinew]{inputenc}
\usepackage{array}
\usepackage{xcolor}
\usepackage{longtable}
\usepackage{multirow}
\usepackage{graphicx}
\usepackage[normalem]{ulem}
\usepackage{algorithmic}
\usepackage{graphicx}
\usepackage{caption}
\usepackage{textcomp}
\usepackage{longtable}
\usepackage{enumitem}
\usepackage{rotating}
\usepackage{hyperref}
\usepackage{amsxtra}
\usepackage{natbib}
\usepackage{makecell}
\usepackage{subfigure}
\usepackage{amsmath,amsthm}
\usepackage[ruled,vlined]{algorithm2e}
\usepackage{float}

\newcolumntype{P}[1]{>{\centering\arraybackslash}p{#1}}
\usepackage{pifont}
\newcommand{\cmark}{\ding{51}}%
\newcommand{\xmark}{\ding{55}}%

\usepackage{soul,color}

\AtBeginDocument{%
  \providecommand\BibTeX{{%
    \normalfont B\kern-0.5em{\scshape i\kern-0.25em b}\kern-0.8em\TeX}}}

\setcopyright{acmcopyright}
\copyrightyear{2021}
\acmYear{2021}
\acmDOI{10.1145/nnnnnnn.nnnnnnn}

\acmJournal{CSUR}

\begin{document}

\title{Federated Learning for Smart Healthcare: A Survey}

\author{Dinh C. Nguyen}
\authornotemark[1]
\email{cdnguyen@deakin.edu.au}
\affiliation{%
  \institution{School of Engineering, Deakin University}
  \city{Waurn Ponds}
  \country{Australia}
  \postcode{3216}
}

\author{Quoc-Viet Pham}
\authornotemark[2]
\email{vietpq@pusan.ac.kr}
\affiliation{%
  \institution{Korean Southeast Center for the 4th Industrial Revolution Leader Education, Pusan National University}
  \city{Busan}
  \country{Korea}
  \postcode{46241}
}

\author{Pubudu N. Pathirana}
\authornotemark[1]
\email{pubudu.pathirana@deakin.edu.au}
\affiliation{%
  \institution{Network Sensing and Biomedical Engineering, School of Engineering, Deakin University}
  \city{Waurn Ponds}
  \country{Australia}
  \postcode{3216}
}

\author{Ming Ding}
\authornotemark[3]
\email{ming.ding@data61.csiro.au}
\affiliation{%
  \institution{Data61, CSIRO}
  \city{Sydney}
  \country{Australia}
    \postcode{2015}
}
\author{Aruna Seneviratne}
\authornotemark[4]
\email{a.seneviratne@unsw.edu.au}
\affiliation{%
  \institution{School of Electrical Engineering and Telecommunications, University of New South Wales (UNSW)}
  \city{Sydney}
  \country{Australia}
    \postcode{2052}
}
\author{Zihuai Lin}
\authornotemark[5]
\email{zihuai.lin@sydney.edu.au}
\affiliation{%
  \institution{Department of Engineering, The University of Sydney}
  \city{Sydney}
  \country{Australia}
  \postcode{2006}
}
\author{Octavia Dobre}
\authornotemark[6]
\email{odobre@mun.ca}
\affiliation{%
  \institution{Faculty of Engineering and Applied Science, Memorial University}
  \country{Canada}
    \postcode{A1B 3X5}
}

\author{Won-Joo Hwang}
\authornotemark[7]
\email{wjhwang@pusan.ac.kr}
\affiliation{%
  \institution{Department of Biomedical Convergence Engineering, Pusan National University}
  \city{Gyeongsangnam}
  \country{Korea}
  \postcode{50612}
}

\renewcommand{\shortauthors}{Dinh C. Nguyen, et al.}


\begin{abstract}
	Recent advances in communication technologies and Internet-of-Medical-Things have transformed smart healthcare enabled by artificial intelligence (AI). Traditionally, AI techniques require centralized data collection and processing that may be infeasible in realistic healthcare scenarios due to the high scalability of modern healthcare networks and growing data privacy concerns. Federated Learning (FL), as an emerging distributed collaborative AI paradigm, is particularly attractive for smart healthcare, by coordinating multiple clients (e.g., hospitals) to perform AI training without sharing raw data. Accordingly, we provide a comprehensive survey on the use of FL in smart healthcare. First, we present the recent advances in FL, the motivations, and the requirements of using FL in smart healthcare. The recent FL designs for smart healthcare are then discussed, ranging from resource-aware FL, secure and privacy-aware FL to incentive FL and personalized FL. Subsequently, we provide a state-of-the-art review on the emerging applications of FL in key healthcare domains, including health data management, remote health monitoring, medical imaging, and COVID-19 detection. Several recent FL-based smart healthcare projects are analyzed, and the key lessons learned from the survey are also highlighted. Finally, we discuss interesting research challenges and possible directions for future FL research in smart healthcare.
\end{abstract}




\maketitle
\input{Introduction.tex}

\input{State-of-Art.tex}
\input{Motivations}
\input{RecentAdvancesFL.tex}
\input{Applications.tex}
\input{Projects.tex}

\input{Challenges.tex}

\input{Conclusion.tex}

\bibliography{Ref}
\bibliographystyle{IEEEtran}

\end{document}

%% file: Introduction.tex
\section{Introduction}
\label{Sec:Introduction}
The revolution in the Internet-of-Medical-Things (IoMT) has transformed the healthcare industry for improving the quality of human life \cite{1}. In the smart healthcare environment, IoMT devices such as wearable sensors are widely used to collect medical data for intelligent data analytics enabled by artificial intelligence (AI) \cite{9200330} to realize a plethora of exciting smart healthcare applications, such as remote health monitoring and disease prediction. For example, AI techniques such as deep learning (DL) have demonstrated their great potential in bio-medical image analytics for the early detection of chronic diseases by handling a large amount of health data to \textcolor{black}{facilitate the provision of healthcare services \cite{2}. }

Traditionally, smart healthcare systems often rely on centralized AI functions located at the cloud or the data center for health data learning and analytics. Given the increasing volumes of health data and the growth of IoMT devices in modern healthcare networks, this centralized solution is not efficient in terms of communication latency due to raw data transmission and cannot achieve high network scalability. Further, the reliance on such a central server or third party for data learning raises critical privacy issues, e.g., user information leakage and data breach \cite{4}. This is particularly true in e-healthcare, where health-related information is highly sensitive and private subject to health regulations such as \textcolor{black}{the} United States Health Insurance Portability and Accountability Act (HIPPA) \cite{6}. Moreover, in the future healthcare systems, such a centralized AI architecture may be no longer suitable because health data are not centrally located, but distributed over a large-scale IoMT network. Therefore, there is an urgent need to go toward distributed AI approaches for enabling scalable and privacy-preserving intelligent healthcare applications at the network edge.

\textcolor{black}{In this context, federated learning (FL) has emerged as a promising solution for realizing cost-effective smart healthcare applications with improved privacy protection \cite{warnat2021swarm, sheller2020federated, kaissis2021end, kaissis2020secure}.} Conceptually, FL is a distributed AI approach which enables the training of high-quality AI models by averaging local updates aggregated from multiple health data clients, e.g., IoMT devices, without the need for direct access to the local data \cite{nguyen2021federated}. This potentially prevents disclosing sensitive user information and user preference, and thus mitigates privacy leakage risks. Moreover, since FL attracts large computation and dataset resources from a number of health data clients to train AI models, the health data training quality, e.g., accuracy, would be significantly improved which might not be achieved by using centralized AI approaches with less data and limited computational capabilities.
\begin{table*}
    \renewcommand{\arraystretch}{1.20}
	\centering
	\caption{{Existing surveys on FL-related topics and our new contributions.}}
	\resizebox{\textwidth}{!}{%
	\begin{tabular}{|p{0.85cm}|p{1.850cm}|p{1.45cm}|p{1.30cm}|p{1.30cm}|p{1.8cm}|p{2.7cm}|p{4.55cm}|}
		\hline
		\centering \multirow{2}{*}{\textbf{Paper}} 	&
		\centering \multirow{2}{*}{\textbf{Key topic}} 	&	
		\multicolumn{4}{c|}{\textbf{Recent Advances in FL}}  	& \multirow{2}{*}{\textbf{Taxonomy}} 
		& \multirow{2}{*}{\textbf{Highlights}} 
		\\ \cline{3-6}
		& & Resource management & Security and privacy & Incentive mechanism & Personalized FL &	&
		\\ \hline
		\multirow{3}{*}{\cite{9}} & \multirow{3}{*}{\makecell{FL concept}} &	\xmark&	\xmark & \xmark & \xmark & None & A discussion of the architectures, algorithms, and data processing methods in FL systems. 
		\\ \hline
		\multirow{2}{*}{\cite{10}} & \multirow{2}{*}{\makecell{FL concept}} &	\cmark&	\cmark&	\xmark&	\xmark & None & A survey of the FL concepts, technologies and associated learning approaches. 
		\\ \hline
		\multirow{2}{*}{\cite{11}} &	\multirow{2}{*}{\makecell{Security and \\privacy in FL}} & \xmark&	\cmark&	\xmark&	\xmark&	None &	A review on the security and privacy issues in FL systems. 
		\\ \hline
		\multirow{2}{*}{\cite{16}} & \multirow{2}{*}{\makecell{FL in edge \\networks}} & \cmark&	\xmark&	\cmark&	\xmark&	None & 	A survey on the integration of FL in mobile edge networks. 
		\\ \hline
		\cite{12} &	\makecell{FL for IoT} &	\cmark&	\cmark&	\cmark&	\xmark &	None & A survey on the use of FL in IoT networks. 
		\\ \hline
		\multirow{3}{*}{\cite{13}} & \multirow{3}{*}{\makecell{FL for IIoT}} & \cmark & \cmark&	\xmark&	\xmark&	The discussion of FL in healthcare is very limited. & A survey on the combination of FL and IIoT, mostly focusing on technical issues in FL implementation.
		\\ \hline
		\multirow{3}{*}{\cite{14}} & \multirow{3}{*}{\makecell{FL for health \\informatics}} & \xmark & \xmark&	\xmark&	\xmark&	The discussion of FL in healthcare is very limited. & A study on the FL architectures and models, with a very short introduction to healthcare. 
		\\ \hline
		\multirow{2}{*}{\cite{15}} & \multirow{2}{*}{\makecell{FL for \\digital health}} & \xmark & \xmark & \xmark & \xmark & The discussion of FL in healthcare is very limited. & A discussion of technical issues and requirements of FL in digital health.
		\\ \hline
		\multirow{2}{*}{\makecell{Our \\work}} &	\multirow{2}{*}{\makecell{FL for smart \\healthcare}} & \cmark &	\cmark &	\cmark&	\cmark & A holistic taxonomy is presented. & A comprehensive survey on the use of FL in smart healthcare, from motivations, requirements to FL designs and applications in a wide range of healthcare domains. 
		\\ \hline
	\end{tabular}
	\label{Table:Comparisons}
	\vspace{-0.1in}
	}
\end{table*}

\subsection{Comparison and Our Contributions}
Driven by the recent advances of FL, many studies have been conducted to survey its related topics, including healthcare. For example, the works in \cite{9} and \cite{10} present the key FL concept and its enabling protocols and technical challenges in FL design and implementation. The survey in \cite{11} discusses the security and privacy issues in FL systems, and described possible solutions for evaluations of malicious threats in FL networks. The integration of FL in mobile edge networks is investigated in \cite{16}, where challenges in FL implementation are explored, such as communication costs, resource allocation, and privacy and security.  Meanwhile, the convergence of FL and Internet-of-Things (IoT) is explored in \cite{12}, by providing a survey on the technical issues in FL designs, such as sparsification, robustness, privacy, and scalability, along with a brief discussion of FL applications in IoT. Moreover, the authors in \cite{13} present an overview of the FL applications in industrial IoT, where the focus is on the discussion of characteristics and fundamentals of FL, while the discussion of FL usage in healthcare is limited. The work in \cite{14} mainly discusses the FL architectures and models with a very brief introduction to the roles of FL in healthcare informatics. Another study in \cite{15} mostly considers technical issues and requirements of using FL in future digital health. \textcolor{black}{However, the latest advances in FL such as resource-aware FL, secure and privacy-enhanced FL, incentive-aware FL, and personalized FL have not been fully explored.} The comparison of the related works and our paper is summarized in Table~\ref{Table:Comparisons}.

\textcolor{black}{Despite these research efforts, there is no existing work to provide a comprehensive survey of the applications of FL in smart healthcare, to the best of our knowledge. Moreover, a holistic taxonomy of the use of FL in emerging healthcare applications is still missing in the open literature. These limitations motivate us to conduct an extensive review of the integration of FL in smart healthcare. Particularly, we first identify the key motivations and highlight the requirements of using FL in smart healthcare. Then, we discover the latest advanced FL designs used for smart healthcare. Subsequently, we provide a state-of-the-art survey on emerging applications of FL in smart healthcare, such as electronic health record (EHR) management, remote health monitoring, medical imaging, and  COVID-19 detection. The lessons learned from the survey are also summarized to provide readers with more insights into the use of FL in smart healthcare. Finally, the research challenges and future directions in FL-smart healthcare are highlighted.} To this end, the key contributions of this article are summarized as follows:
\begin{enumerate}
	\item We present a state-of-the-art survey on the use of FL in smart healthcare, beginning with an introduction to the FL concept and  discussions of the motivations as well as the technical requirements for the utilization of FL  smart healthcare. 
	\item	We present the recently advanced FL designs that would be useful to federated smart healthcare applications, including resource-aware FL, secure and privacy-enhanced FL, incentive-aware FL, and personalized FL. 
	\item	We provide an updated review on the key applications of FL in smart healthcare via a wide range of key domains, namely federated EHRs management, federated remote health monitoring, federated medical imaging, and federated COVID-19 detection. The emerging real-world projects related to FL-healthcare use cases are provided, and the key lessons learned from the survey are also highlighted. 
	\item	Finally, we highlight interesting challenges and discuss future directions in FL-smart healthcare.
\end{enumerate}

\subsection{Structure of The Survey}
The remainder of the article is organized as follows. Section~\ref{Sec:State-of-Art} introduces the key principle of FL and describes the key FL types used in smart healthcare. The key motivations and technical requirements of the use of FL in smart healthcare are explained in Section~\ref{Sec:Motivations}. Subsequently, we present the advanced FL designs that are useful to federated smart healthcare in Section~\ref {Sec:RecentAdvancesFL}. In Section~\ref{Sec:Applications}, we present a state-of-the-art review on the emerging applications of FL in smart healthcare, namely federated EHRs management, federated remote health monitoring, federated medical imaging, and federated COVID-19 detection. The real-world projects of FL implementation for smart healthcare via some practical use cases are highlighted in Section~\ref{Sec:Projects}. We discuss the key challenges and future directions related to FL-Healthcare research in Section~\ref{Sec:Challenges_Future-Directions} and Section~\ref{Sec:Conclusion} concludes the article.

%% file: State-of-Art.tex
\section{FL for Healthcare: Key Principle and Categories}
\label{Sec:State-of-Art}
In this section, we present the key principle of FL and describe the key FL types used in smart healthcare.
\begin{figure*}
	\centering
	\includegraphics[width=0.97\linewidth]{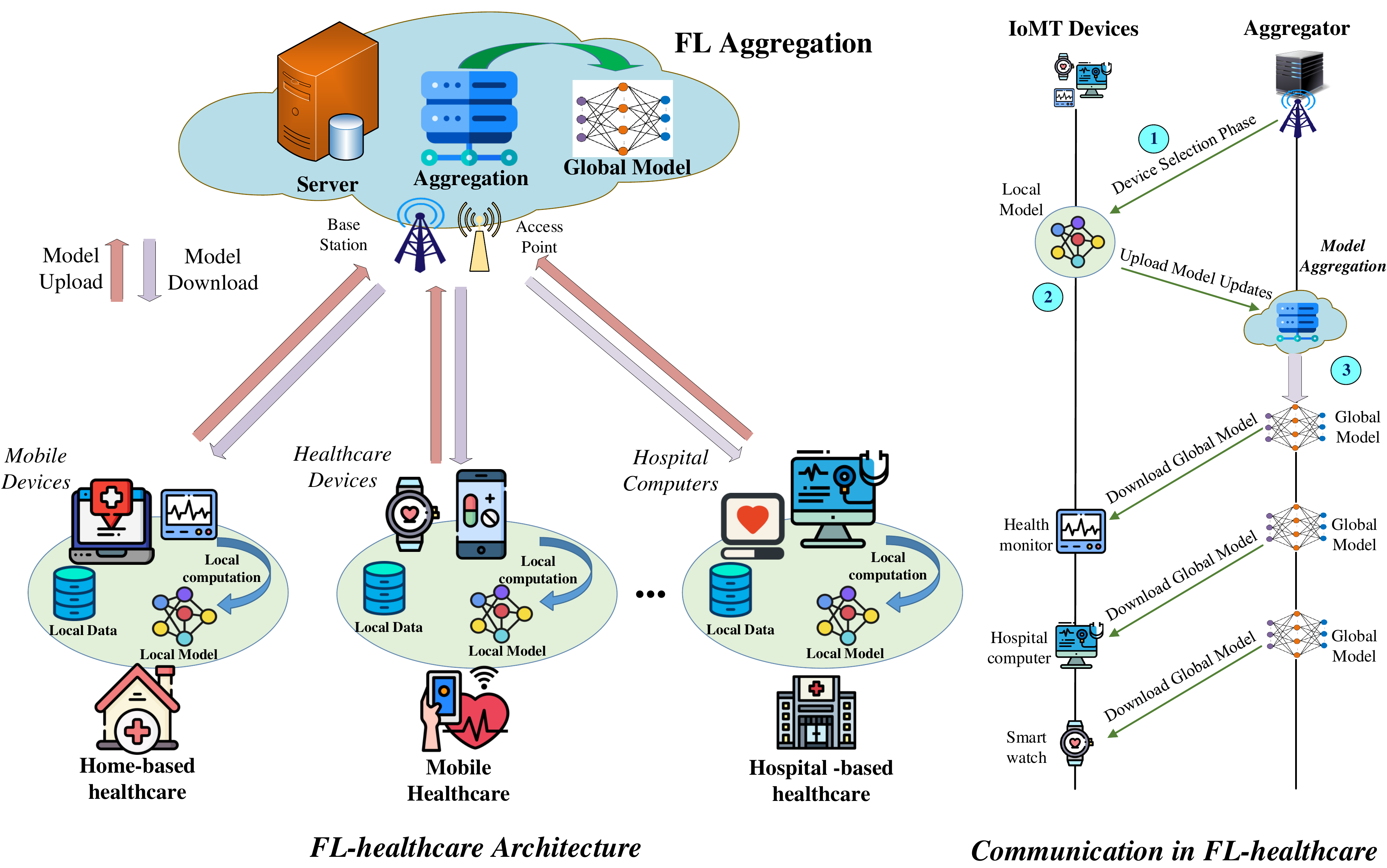}
	\caption{{\color{black}The network architecture and communication process for FL-smart healthcare. }}
	\label{Fig:FL_Concept}
	\vspace{-0.1in}
\end{figure*}
\subsection{Key Principle}
As shown in Fig.~\ref{Fig:FL_Concept}, the generic FL-smart healthcare process includes the following key steps.
\begin{enumerate}
	\item 	\textit{System Initialization and Client Selection:} The aggregation server selects a healthcare analytic task, e.g., automatic medical imaging or human motion detection, along with model requirements such as task classification or task prediction, and learning parameters \textcolor{black}{such as neural node numbers and learning rates}. Moreover, the server selects a subset of clients that should be involved in the FL process.
	\item	\textit{Distributed Local Training and Updates:} \textcolor{black}{Once the subset of the learning clients is determined, the server sends an initial model including an initial global gradient to the clients to trigger the distributed training. In every communication round, each client trains a local model using its own dataset and calculates its model update, e.g., the gradient in neural networks. Then, each client uploads its model update to the server for aggregation.}
	\item	\textit{Model Aggregation and Download:} \textcolor{black}{After receiving all updates from the selected clients, the server aggregates them by using an aggregation method. For example, we can use the model averaging approach in the Federated Averaging (FedAvg) algorithm proposed by Google \cite{18}, where the gradient parameters of local models are averaged element-wise with weights proportional to the sizes of the client datasets.} Subsequently, the server calculates a new version of the global model and broadcasts it to all clients \textcolor{black}{as the basis for further local model updates in the next learning round.} The FL process is iterated until the global loss function converges or the desired accuracy is achieved.
\end{enumerate}

\subsection{Types of FL for Smart Healthcare}
\begin{figure*}[t]
	\centering
	\includegraphics[width=0.98\linewidth]{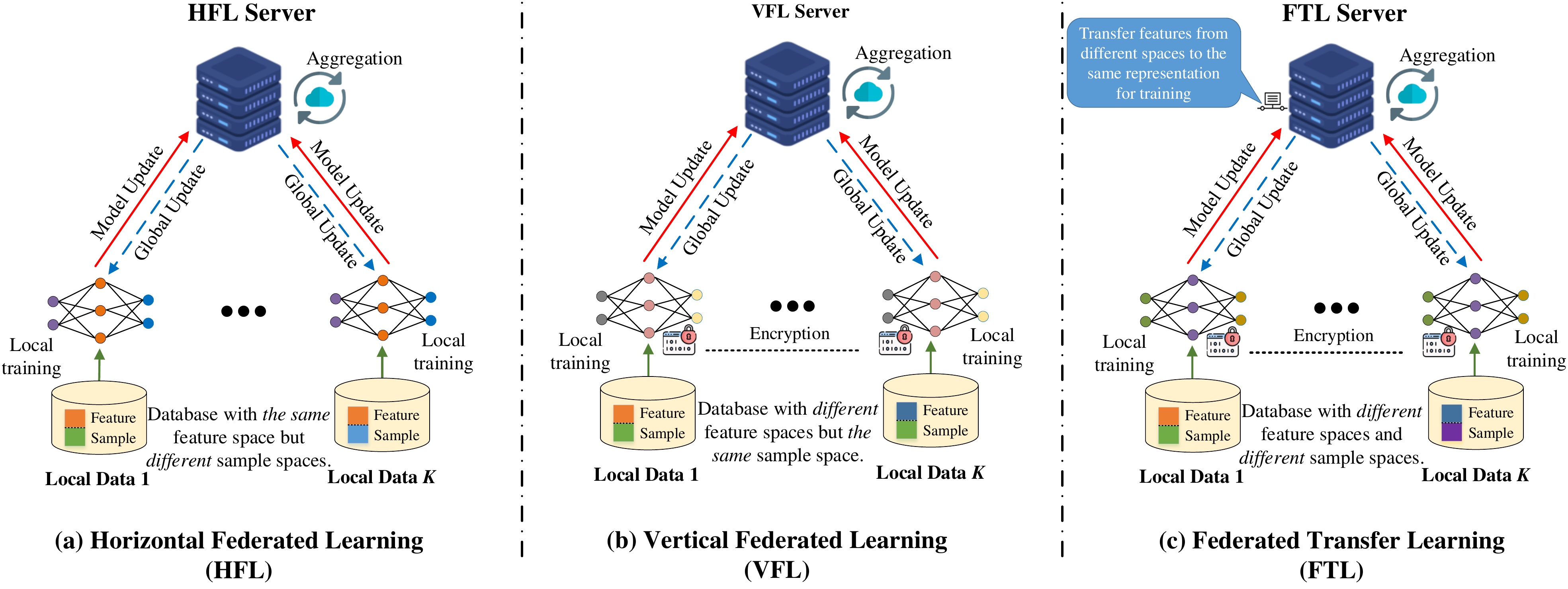}
	\caption{{Categories of FL used in smart healthcare. }}
	\label{Fig:FL_Classifications}
	\vspace{-0.1in}
\end{figure*}

Reviewing the recent advances of FL algorithms used in smart healthcare, we categorize FL into three types as illustrated in Fig.~\ref{Fig:FL_Classifications}. 
\begin{itemize}
	\item \textit{Horizontal FL (HFL):} In HFL, the healthcare clients can participate in training a shared global model using their datasets which own the same feature space while having different sample spaces, as shown in Fig.~\ref{Fig:FL_Classifications}(a). In this regard, local FL participants can adopt the same AI model (e.g., neural network-NN) for training their datasets. Subsequently, the server will combine the local updates transmitted from local participants to build a global update without the need for direct access to local data \cite{19}. \textcolor{black}{An HFL example in smart healthcare can be the detection of speech disorders where multiple users speak the same sentence (feature space) with different voices (sample space) on their smartphones and then the local speaking updates are averaged by a parameter server to create a global model for speech recognition.}
	\item \textit{Vertical FL (VFL):} \textcolor{black}{VHL works on the federated training of health datasets which have the same sample space with different data feature spaces}, as illustrated in Fig.~\ref{Fig:FL_Classifications}(b). Particularly, to address the issue of data sample overlapping at distributed clients, solutions based on entity alignment can be employed by integrating with encryption techniques during the local training \cite{20}. An example of VFL in IMoT applications can be the shared learning model among entities in a smart healthcare environment, e.g., hospitals and an insurance company. In this context, a hospital and an insurance company (different data feature) which serve patients (same sample space) can join a VFL process to cooperatively train an AI model using their datasets, e.g., historical medical records at hospitals and healthcare costs at the insurance company for intelligent healthcare decision making. 
	\item \textit{Federated Transfer Learning (FTL):}  \textcolor{black}{Unlike VFL systems, FTL is provided to handle datasets with different sample spaces and different feature spaces}, as shown in Fig.~\ref{Fig:FL_Classifications}(c). By using a transfer learning method, feature values are calculated from different feature spaces to the same representation which is exploited to train local datasets \cite{21}. Encryption techniques such as random masks are also useful to provide further privacy protection during the gradient exchange between clients and the server \cite{22}. In smart healthcare, FTL can support disease diagnosis by collaborating countries with multiple hospitals that have different patients (sample space) with different therapeutic programs (feature space). In this way, FTL can enrich the shared AI model output for improving the accuracy of diagnosis.
\end{itemize}

%% file: Motivations.tex
\section{Motivations and Requirements of Using FL in Smart Healthcare}
\label{Sec:Motivations}
In this section, we explain the key motivations and describe the technical requirements of the use of FL in smart healthcare in details. 
\subsection{Motivations}
To highlight the motivations, we first identify the limitations of current healthcare systems, and then discuss the key benefits that FL can offer for healthcare. 
\subsubsection{Limitations of Current Smart Healthcare Systems}
\begin{itemize}
	\item \textit{Privacy Concerns:} As briefly discussed in the previous sections, the use of traditional AI-based approaches for realizing smart healthcare require open data sharing with cloud or data centres which makes health information vulnerable to privacy attacks. Indeed, adversaries can gain unauthorized access to the AI training centres for data retrieval, or third parties such as cloud providers can gain the control over the data and modify data patterns without the consent of users \cite{23}. These bottlenecks would result in serious issues of data leakage and the user confidentiality \textcolor{black}{can be compromised}. Although the cloud servers with powerful computation capabilities can provide efficient data training and analytics, such a centralized AI-based smart healthcare solution comes at \textcolor{black}{a heavy cost of privacy risks \cite{24}. }
	\textcolor{black}{\item \textit{Lack of Datasets at Medical Sites:} In realistic healthcare systems, the dataset of a single medical site (e.g., a clinical lab) may be insufficient to run the AI model which can  prevent a proper health data training \cite{25}. This makes AI-based smart healthcare solutions inefficient and manual data analyses need to be taken, but this incurs long data processing delays. A possible solution is to exchange the data between medical sites to support the data training, but given the institutional policies and growing user privacy concerns, it is not easy to obtain data from other sites to train the AI model \cite{26}. Hence, how to solve the issues of dataset shortage  is of paramount importance for smart healthcare system designs. }
	\item \textit{Limited Health Data Training Performance:} Due to the lack of datasets, the training at an single medical site cannot achieve the desired degree of accuracy, e.g., disease classification accuracy. The reasons behind this observation can come from the imbalance of data features and the insufficient data sizes. One can use data augmentation techniques such as generative adversarial networks (GANs) \cite{27} to solve these issues, but it may still not have good diversity to build a comprehensive dataset for efficient data training. This is also one of the most critical challenge in applying AI in healthcare, where the training becomes more difficult due to the limited datasets. 
	\item \textit{High Costs in Health Data Training:} In the traditional AI-based smart healthcare systems, the offloading of health data to the cloud for execution incurs excessive network latency \cite{507}, especially when medical data often have large sizes (e.g., audio, images). Moreover, the health data transfer also consumes much network bandwidth, which is likely to cause network congestion when the number of devices increases. The offloading process also requires transmit power of medical devices which in turn poses new challenges on battery and hardware designs on devices. 
\end{itemize}
\subsubsection{Benefits of FL in Smart Healthcare}
Based on the innovative operational concept, FL is able to bring many attractive benefits to advance smart healthcare, as explained below: 
\begin{itemize}
	\item \textit{Data Privacy Improvement:} In the FL-based smart healthcare system, only the local updates such as model gradients are required by the central server for the AI training, while the local health data are kept at local medical sites and devices. This would reduce the risks of the leakage of sensitive user information to the external third-party, and thus providing a higher degree of user privacy \cite{29}. Following the increasingly stringent health data privacy protection legislation, the capability of preserving health user information of FL is important for building sustainable and safe smart healthcare systems\footnote{\textcolor{black}{However, it should be noted that FL cannot fully address the privacy problem in smart healthcare \cite{ma2020safeguarding}. Dedicated privacy protection mechanisms thus need to be designed to enhance FL in healthcare networks.}}. 
	\item \textit{Reasonable Trade-off between Accuracy and Utility:} \textcolor{black}{Compared with conventional centralized learning, FL is able to offer a reasonable trade-off between accuracy and utility along with privacy enhancement. Moreover, FL training retains the model generalizability at the cost of nominal accuracy loss.} In return, FL can enhance the scalability of the smart healthcare system thanks to its distributed learning feature.
	\item \textit{Low-cost Health Data Training:} By avoiding the offloading of huge data volumes to the server, FL can help reduce significantly communication costs, e.g., latency and transmit power, consumed by raw data transmission, as the model gradients generally have much smaller sizes compared to their actual datasets \cite{31}. As a result, FL also save much network bandwidth and mitigate possibility of network congestion in massive healthcare networks. 
\end{itemize}
\subsection{Requirements}
To realize the full potential of FL in smart healthcare, several requirements should be met as highlighted below:
\subsubsection{Trusted Server}
One of the most important entities in FL is the central server that is used to aggregate local model gradients to build the global model in each communication round. Although the FL concept can provide privacy protection by allowing users to keep their data at local sites during the training, it has been proven that the model updates might still contain health user-related information such as data features and image resolution that can be re-constructed by the curious global server \cite{FLchain}. As a result,  user privacy can be put at risks during the training, which also makes FL vulnerable and discourages medical sites from joining the collaborative training. Based on this fact, a fundamental requirement to ensure reliable FL operations in smart healthcare is to build a trusted server for the data training coordination and model aggregation. The computation services provided by the global server need to ensure a transparent and reliable model aggregation under the agreement between the service provider and healthcare organizations such as local hospitals. This is particularly necessary for the smart healthcare domain, where health data are highly sensitive and the data computation outside the data sources must be trusted to provide reliable FL-based healthcare. To further build trust for the server, recent research efforts have been put to develop new solutions, such as building decentralized and trusted servers  enabled by blockchain \cite{Blockchain1} or providing secure aggregation methods, which will be detailed in the following section. 
\subsubsection{Reliable Client-Server Communications}
Another important requirement for FL-based smart healthcare is the reliable communication between local clients and the global server. The exchange of local model updates to the server may be risky due to external threats \cite{33}. Indeed, an adversary can deploy data attacks to the communication channels established by the clients and the server to steal the updated  information which can interrupt the model computation at the global server due to the insufficient local updates to build a global model. Further, an attack can make attempt to modify or change the local updates which leads to aggregation bias at the server. All of these security concerns require the provision of safe and reliable client-server communications before deploying the FL functions in the smart healthcare system. This also builds up trust for health users in joining the FL process to solve collaborative health tasks  such as federated medical image analysis.  
\subsubsection{Computational Capability for Local Training at Health Clients}
In FL-based mobile smart healthcare where mobile medical devices participate in the federated data training, their computational capability is a key concern. In fact, to realize federated smart healthcare, one needs to join in the multiple communication rounds to achieve a desired training performance. In this regard, certain medical devices such as lightweight smart watches may be unable to join the training in the long run due to their limited computation ability and less energy resources \cite{34}. Without the involvement of multiple devices in the training, the FL concept becomes inefficient in smart healthcare, where the contributed computation from different devices are highly important to improve health data training. Hence, how to build computation-accelerated hardware for health devices is essential to build FL-based smart healthcare ecosystems.  
\subsubsection{Available Dataset at Health Clients}
To obtain the desirable training performance in FL-based smart healthcare, the availability of datasets at clients is needed. One device or medical site is required to construct its own dataset based on its working environment, e.g., a smartphone can collect human motion data of patients under its working area. Each participating client can also prepare its own data features based on its collected dataset via data-driven techniques such as feature extraction, for its own local training. In this context, one of the key concerns in dataset preparation is the non-independent and identically distributed (non-IID) issue which potentially makes the FL training highly divergent in the data training. Several solutions to cope with non-IID issues thus need to be developed, e.g., creating an additional subset of datasets to allocate fairly among clients \cite{35}, aiming to ensure efficient data training in FL-based smart healthcare. 

\textcolor{black}{For designing an efficient FL system, one needs to take its related operational metrics into account, such as the size of the deep learning models, the size of the local data, convergence time, and required accuracy. These performance metrics depend on the device, network conditions, and calculation speeds of both clients and the server. Such quantitative evaluations can be found in a recent work \cite{bonawitz2019towards}. }

%% file: RecentAdvancesFL.tex
\section{Advanced FL Designs for Smart Healthcare}
\label{Sec:RecentAdvancesFL}



This section presents the advanced FL designs for smart healthcare, including resource-aware FL, secure FL, privacy-enhanced FL, incentive-aware FL, and personalized FL.

\subsection{Resource-aware FL}
\label{Subsec:Resource-awareFL}
Collaborative FL models are built based on local model updates of IoMT devices in the uplink, communications with the aggregation server, and the global model broadcasting in the downlink. In this regard, resource management plays an important role in improving the performance of FL-enabled healthcare applications.  

A scheduling problem is investigated in \cite{xu2021online} to minimize the total training time by deciding a set of IoMT devices. However, finding the optimal solution to the non-integer scheduling problem with massive devices is very challenging. To overcome this challenge, along with one caused by unknown channel state information between IoMT devices and the aggregation server, the multi-armed bandit (MAB) theory is adopted to find the solution. The experimental results show that the MAB approach results in significantly low training loss compared with several benchmarks as well as outperformance in terms of the training latency. A similar scheduling problem using the MAB theory is studied in \cite{xia2020multi}. In particular, two scenarios are considered: 1) when all IoMT devices are available and the data distribution is IID, and 2) when IoMT devices are not always available and the data distribution is non-IID. This work concludes an interesting observation on the relationship between the number of IoMT devices and convergence rate.
Given the importance of resource management, joint optimization of device scheduling and resource allocation are proposed in the literature. For example, the work in \cite{luo2020hfel} proposes a hierarchical federated edge learning framework, in which the model aggregation is partially done by immediate hospitals. Under this framework, a joint resource allocation and device association problem is formulated and then solved by an iterative algorithm. Another joint optimization of device scheduling and resource allocation framework is investigated in \cite{shi2021joint}. In particular, a lower bound on the training time is derived and a greedy algorithm is designed to jointly optimize bandwidth allocation and device scheduling. In \cite{yang2020scheduling}, the concept of distributed coordinate descent from the optimization perspective is leveraged to analyze the effects of scheduling policies on the convergence rate of FL algorithm. Such analyses are important in scenarios where the aggregation server needs to connect with massive IoMT devices. Three scheduling schemes are examined in \cite{yang2020scheduling}, including random scheduling, round-robin, and proportional fair, showing that the convergence rate of these scheduling policies is largely dependent on the signal quality threshold. 

A promising direction towards resource-aware FL for smart healthcare is to optimize joint computing and radio resources. The work in \cite{nguyen2020toward} to considers a joint computation and communication resource allocation framework for multiple FL services, which is motivated by the success of AI-enabled multiple services and applications deployed simultaneously at mobile devices and the network edge (e.g., virtual reality, video streaming, and healthcare apps). The multiple FL service problem is then formulated to minimize the total training overhead in terms of energy consumption and completion time \cite{pham2021uav}. Two algorithms are designed, including a centralized algorithm based on the block coordinate descent method and a decentralized algorithm based on the Jacobi-Proximal alternating direction method of multipliers (ADMM) approach. 
Multiple FL services are also investigated in a recent work \cite{xu2021bandwidth}. Different from \cite{nguyen2020toward} where diffident FL services running on the same devices, \cite{xu2021bandwidth} considers that each device runs only one FL service and multiple FL services are available at multiple IoMT devices in the network. A bandwidth allocation scheme is designed to allocate bandwidth resources among clients running the same FL service and among FL services. A distributed algorithm is firstly proposed to guarantee fairness among FL services, following by an auction-based game-theoretic approach to balance fairness and performance. 

\subsection{Secure FL}
\label{Subsec:SecureFL}

\begin{figure*}[t]
	\centering
	\includegraphics[width=0.65\linewidth]{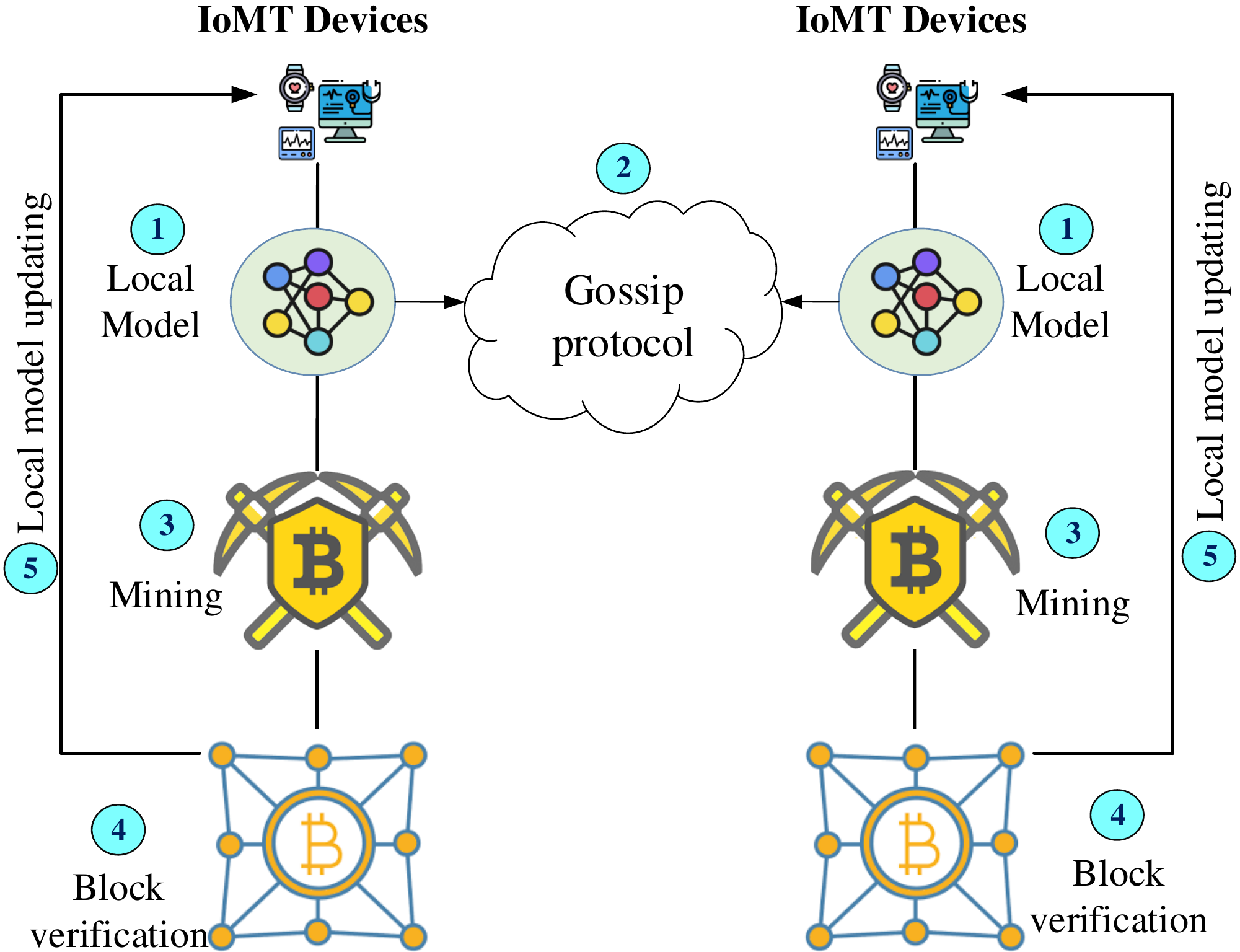}
	\caption{Illustration of a single communication round of blockchain-enabled FL healthcare systems.}
	\label{Fig:FL_BC_Concept}
	\vspace{-0.1in}
\end{figure*}
\textcolor{black}{It is important to study secure solutions for FL-enabled healthcare applications} since there are potential security attacks in FL systems such as \textcolor{black}{poisoning attacks, inference, backdoor attacks, malicious server, communication bottlenecks, and free-riding attacks \cite{tolpegin2020data,fang2020local,wang2020attack}}. These security attacks can be caused by numerous sources, including communication protocols, data manipulation, and aggregation algorithm \cite{mothukuri2021survey}. In the context of smart healthcare, a number of solution approaches have been investigated over the last few years. To prevent unreliable updates from untrusted devices, a new concept called reputation is introduced in the context of FL-enabled healthcare systems \cite{kang2020reliable}. Such reliable device selection plays an important role in mitigating several security attacks. For example, a malicious IoMT device may inject poisonous data into its local data, and thus updating the fall local model may degrade the accuracy of the global model. Reliable device selection is also important in scenarios, where local FL models are trained using low-quality and noiseless data. 

\textcolor{black}{Decentralized FL is a promising solution for secure federated smart healthcare to address the problem of untrusted parameter servers in centralized FL. The implementations of decentralized FL are typically based on  gossip, consensus, and diffusion approaches. For example, a decentralized FL scheme is proposed in \cite{hu2019decentralized} using a segmented gossip aggregation method to enhance the performance of federated data training. Each data client can act as a worker that stochastically selects a few neighbour workers to transmit the model segment in every training iteration, aiming to optimize the utilization of bandwidth capacities of all clients. The simulation results show its superior performance in terms of reduced training time in practical network topology and bandwidth settings with only slight accuracy degradation, compared to centralized FL. A decentralized FL design for smart healthcare is proposed in \cite{lu2020decentralized} over a graph. In this case, the FL algorithm allows local clients to perform local updates for several iterations and then enables peer-to-peer communications among health clients such as distributed hospitals. In such a way, the delay of parameter exchange and communications among clients can be reduced since there is no need to transfer the models to the remote central server.}

However, designing reliable user selection in smart healthcare systems faces several challenges such as no universal metric for reliability evaluation, no universal selection scheme, and no real-time user monitoring methods. To address these challenges, blockchain is adopted in \cite{kang2020reliable} to manage FL users' reputation. \textcolor{black}{Particularly, the integration of blockchain into federated settings decentralizes the FL that allows for the elimination of a single central server in the model aggregation \cite{passerat2020blockchain}. The blockchain can coordinate the calculation of global model via the block consensus among participants in a peer-to-peer manner. The use of blockchain to prevent untrustworthy server and external attacks in decentralized FL-enabled healthcare systems is also studied in \cite{li2021blockchain}. Here, blockchain helps avoid the need for a centralized server (i.e., located at hospitals or clinics), whereas in each global communication round, IoMT devices compete with each other to mine a block and then append a new block to their local ledgers.}

The illustration of a single communication round of blockchain-enabled FL healthcare systems is shown in Fig.~\ref{Fig:FL_BC_Concept}. In sum, the procedure in a global round of such a blockchain FL system consists of five main steps as the following.
\begin{enumerate}
    \item \textit{Local training}: Each IoMT device trains the local FL model using its local data.
    
    \item \textit{Model broadcasting and verification}: Each device adds its digital signature to the model and broadcasts the model to the other IoMT devices via some gossip protocols. The transaction of a device is then verified by all other IoMT devices in the network.
    
    \item \textit{Mining}: Upon receiving the local models from the other devices, each IoMT device tries to mind the current block.
    
    \item \textit{Block validation}: the current block, if verified, is added to local ledgers of IoMT devices. 
    
    \item \textit{Local model updating}: Upon receiving the verified, each device updates its local model and starts a new communication round.
\end{enumerate}
The upper bound of the loss function is also analyzed in \cite{li2021blockchain}, which is shown to be a function of the communication rounds and other factors such as the training time of each local round, mining time, learning rate, data distribution. and computation time. Based on that derived upper bound of the loss function, the impact of the number of lazy devices is further evaluated. More lazy IoMT devices decrease the learning performance, e.g., no lazy client can achieve an accuracy of 85.53\% while the accuracy is only 78.80\% when 30\% clients are lazy. The application of blockchain and game theory for the design of a reliable FL algorithm is studied in \cite{kang2019incentive}. Particularly, the contract theory is used to encourage the participation of highly reputed IoMT devices and a public blockchain approach is employed to manage the reputation updates of FL devices in a decentralized and secure manner. 

An important problem in securing FL is secure aggregation; that is, the multiparty aggregation at the central server is not revealed by external attackers. According to \cite{bonawitz2016practical}, secure aggregation approaches can be classified into secure communication protocols, dining cryptographers (DC) based secure aggregation, homomorphic threshold encryption, and pairwise masking. It is also suggested that a secure aggregation method needs to satisfy the following conditions:
\begin{enumerate}
    \item It can work with high-dimensional updates from users, 
    
    \item It is communication-efficient even with massive users, 
    
    \item It is robust to the user participant and unavailability (also known as user dropping-out), 
    
    \item It can provide security guarantees under unfavorable environments such as unauthenticated transmission channels and resource-constrained edge nodes.
\end{enumerate}
A seminal work on secure FL aggregation is conducted in \cite{bonawitz2016practical}. To address the challenge caused by user dropping-out, FL users are required to share a part of their Diffie-Hellman keys with all the other users, and thus the pairwise can be completely recovered if some FL users are no longer active. Such a problem typically happens in the context of healthcare applications; for example, some patients only visit the hospital for a very short period of time. A double-masking scheme is also leveraged to protect the user privacy, i.e., the server cannot know the user's masks. To provide more practical solutions, \cite{bonawitz2016practical} also proposes two variants to balance the security guarantee level and communication efficiency. Recently, a secure aggregation method, namely Turbo-Aggregate, is proposed in \cite{so2021turbo}. More specifically, the Turbo-Aggregate method comprises three main components, including multi-group circular strategy, additive secret sharing, and Lagrange coding. The first component is to divide the entire set of FL users into multiple groups and each group needs to send the aggregated model from itself and other groups to the next group. The second component is to enhance the user privacy via additive secret sharing, and the third component is to increase the robustness against user unavailability. There are three main advantages of Turbo-Aggregate: 1) the aggregation overhead is significantly reduced from $\mathcal{O}(N^2)$ to $\mathcal{O}(N\log(N))$ with $N$ being the number of IoMT devices joining int the FL process, 2) up to 50\% dropping-out rate can be tolerated by Turbo-Aggregate, and 3) the total running time is multiple times faster than the benchmarks (i.e., up to 40 times reported by the experiments). However, such advantages are only evaluated via numerical results and thus experiments on the real health datasets can be examined by using the secure aggregation methods proposed so far. \textcolor{black}{A recent promising solution for model aggregation is called over-the-air aggregation, which exploits the properties of wireless channels to wirelessly aggregate local updates \cite{zhu2021one}.} The over the air aggregation can be extended to take into account the security and privacy aspects of healthcare applications. 

\subsection{Privacy-enhanced FL}
\label{Subsec:Privacy-enhancedFL}
Despite the great potential in improving user data privacy, FL also has its own privacy concerns such as membership attacks, unintentional data leakage, generative adversarial network based inference attacks \cite{mothukuri2021survey}. 
For example, based on the local model updates from an IoMT device, the attacker can infer and recover the local data using the construction attacks \cite{zhu2020deep} and/or predict the existence of a data sample in the local training dataset such as blood type, disease name, gender information, and any other private information, which can be done via membership inference attacks \cite{nasr2018machine}.
This motivates the development of advanced solutions of privacy-enhanced FL designs for smart healthcare applications. 

Differential privacy is a well-known concept that can be adopted to enhance the user privacy in wireless and mobile network. Motivated by this advantage, a number of research works have been devoted to study differentially private FL systems, for general domains as well as smart healthcare services. The work in \cite{wu2021incentivizing} first considers that artificial noise can be added to the local dataset of IoMT devices to protect the user privacy. Then, a multi-dimensional cost problem is studied to design an incentive mechanism. Three types of costs are taken into account to evaluate the contribution of each IoMT device, which are computation cost, communication cost, and privacy cost. Experimental results showed that the proposed three-dimensional contract-based incentive approach with differential privacy can obtain lower training loss and higher training accuracy compared with the seminal FL. There are several works focusing on applying the differential privacy concept to federated health services. For example, the work in \cite{malekzadeh2021dopamine} proposes a method to balance the differential privacy guarantee and deep learning accuracy. In particular, the differentially-private stochastic gradient descent method is employed at the distributed healthcare datasets and a secure aggregation step using momorphic encryption is leveraged. The proposed method is then tested in the DR dataset\footnote{The dataset is available at https://www.kaggle.com/c/aptos2019-blindness-detection/data.} to detect diabetic retinopathy, and SqueezeNet is used as the main deep architecture. Experimental results show that the performance of the proposed method is close the that of the centralized approach, while significantly outperforming several benchmarks such as FL with parallel differential privacy, where different hospitals apply differential privacy on their datasets individually and FL with centralized differential privacy, where a trusted entity is required to apply differential privacy for all the patients. Another differently private FL for healthcare applications is investigated in \cite{38}. Two learning tasks are considered which include the prediction of adverse drug reaction and mortality rate. A real health dataset with over 1 million training samples is tested, which contains many personally private information such as diagnosis results, prescription fills, and admission records. A promising observation from the experiments is that the higher the privacy level is, the lower the training accuracy is. As a result, more research works should be investigated in the future to improve both learning performance and data privacy. 
However, such differently private FL works in \cite{malekzadeh2021dopamine} and \cite{38} are promising as their FL frameworks can avoid the need for transmitting a large amount of data from distributed health organizations to a central entity and also project the patient data against potential privacy attacks. 

In order to further enhance the performance of FL-enabled healthcare services, joint optimization of privacy-preserving and resource management is of importance \cite{36}. The authors in \cite{kerkouche2021privacy} discuss that even the raw data is not needed to be sent to the central server, exchanging model updates from massive IoMT devices is not bandwidth-efficient. Accordingly, a bandwidth-efficient FL scheme is proposed in \cite{kerkouche2021privacy} with privacy guarantee. More specifically, only the sign of updated values is sent to the aggregation server, which then scales the updates and does the aggregation to update the global model. The differential privacy concept is also used to protect every data sample of all the hospitals, but at the same time it ensures that the hospital cannot infer the global model broadcast by the aggregation server. An EHR dataset is used to examine the performance of the proposed scheme, namely FL-SIGN-DP, in terms of in-hospital mortality rate. The FL-SIGN-DP scheme is also compared with a number of existing alternative schemes such as centralized learning, FL-SIGN (i.e., no differential privacy), standard FL, and standard FL with differential privacy. The FL-SIGN-DP is shown to consume a very small amount of bandwidth, i.e., FL-SIGN-DP sends only 1.76 Mb while the standard FL sends up to 56.48 Mb, while FL-SIGN-DP can maintain a reasonable level of privacy protection. 

\subsection{Incentive-aware FL}
\label{Subsec:Incentive-awareFL}
Conventional FL approaches require all IoMT devices to share local model updates with the aggregation server; however, it is not always available in practice. It is since the IoMT devices are usually limited in terms of computing resources, radio bandwidth, and privacy concerns for personal data and server trustworthiness, thus having no willingness to share their models. To incentivize the participation of more FL users and improve the performance of FL-enabled healthcare scenarios, incentive-aware FL solutions are necessary. According to a recent survey \cite{zhan2021survey}, incentive mechanisms designed for FL can be categorized by different aspects, including the device's data contribution, device reputation, and resource allocation. 
\begin{itemize}
    \item \textbf{Data contribution} is evaluated via two important metrics: data quantity and data quality. While data quantity is typically assessed using the Shapely value, data quantity refers to the size of the local model updates and training samples.
    
    \item \textbf{Device Reputation} is an important metric in designing incentive algorithms for FL. \textcolor{black}{In general, reputation reflects the extent that FL users can provide high-quality data for model training and reliable local updates. }
    
    \item \textbf{Resource Allocation} is an important phase of any incentive scheme as computation and communication resources need to be appropriately allocated to FL users so as to improve the FL performance. 
\end{itemize}

Game theory, originally derived from economics and business studies, is an excellent tool for designing incentive mechanisms in wireless and mobile network over the last decade. Motivated by this excellence, several game-theoretic incentive strategies have been investigated for FL-based healthcare systems. For example, the work in \cite{31} leverages the Stackelberg game to incentivize user participation in the FL learning process. In such a game, each IoMT device first receives an offer from the aggregation server located at the hospital and then decides whether or not to participate in the FL learning process. After that, the aggregation server updates the incentive strategy so as to maximize its utility, which can be defined based on the number of global communication rounds and the target learning accuracy \cite{31}. These steps of IoMT devices and the aggregation server are repeated until a desirable performance is obtained. Such a Stackelberg game is promising and can be much extended in the context of smart healthcare. For example, a dynamic Stackelberg game can be studied to consider time-varying computing resources and channel connections between the IoMT devices and the aggregation server located at the hospital \cite{sarikaya2019motivating}. A notable limitation of Stackelberg game approaches is the requirement of the full knowledge of FL devices' contributions. To overcome this limitation, the work in \cite{zhan2020learning} develops a deep reinforcement learning (DRL) method to help the aggregation server to determine the award strategy and the devices to decide their local training schemes. The main advantage of the proposed DRL approach is that the historical payment strategies can be utilized to deal with future scenarios, e.g., more patients/relatives will be interested in joining the FL learning process to receive valuable clinical information from the hospitals and/or insurance companies.

In the context of cross-silo FL, where different hospital/insurance companies try to collaboratively build a global model built at the third-party aggregation server, motivating different hospitals to contribute their resources to the global model so as to maximize the social welfare is an open problem. The work in \cite{tangincentive} studies this problem in order to answer two key questions: how does each health organization allocate its computing resources for local training and how to compensate the local model of each health organization by other organizations computing resources. \textcolor{black}{To overcome the difficulty of solving nonconvex problems,} a distributed algorithm is investigated by using the non-cooperative game. The proposed game is proved to be not only Nash-equilibrium but also has nice properties such as social welfare, individual rationality, and no need for a third-party payment entity. Recently, the work in \cite{zhao2021efficient} proposes an efficient method to evaluate the contribution of data contributors such as wearable computing devices, smart phones, and health handsets. The proposed method in \cite{zhao2021efficient} is to overcome a critical challenge of conventional incentive FL designs using the Shapley value that the computing cost increases exponentially with the number of IoMT devices and the dimensionality of data features. Exploited the success of DRL in long-term evaluation, the contribution of data contributors is efficiently assessed via a DRL-based approach, showing its superior performance to the conventional incentive mechanism.  

\subsection{Personalized FL}
\label{Subsec:PersonalizedFL}
To provide personalized services to IoMT users, the conventional FL faces several challenges. A very first challenge is that the global model only captures the statistical characteristics of different IoMT devices but is hard to provide distinct personal styles. For example, when learning to predict the disease, the same body weight and height from different people may have different meanings due to personal living environments and various external factors. Another challenge is heterogeneous computing resources and network conditions of different FL devices. Motivated by the importance of personalized FL models, recent years have witnessed several advances in personalized FL designs \textcolor{black}{for} smart healthcare.

\begin{figure*}[t]
	\centering
	\includegraphics[width=0.8\linewidth]{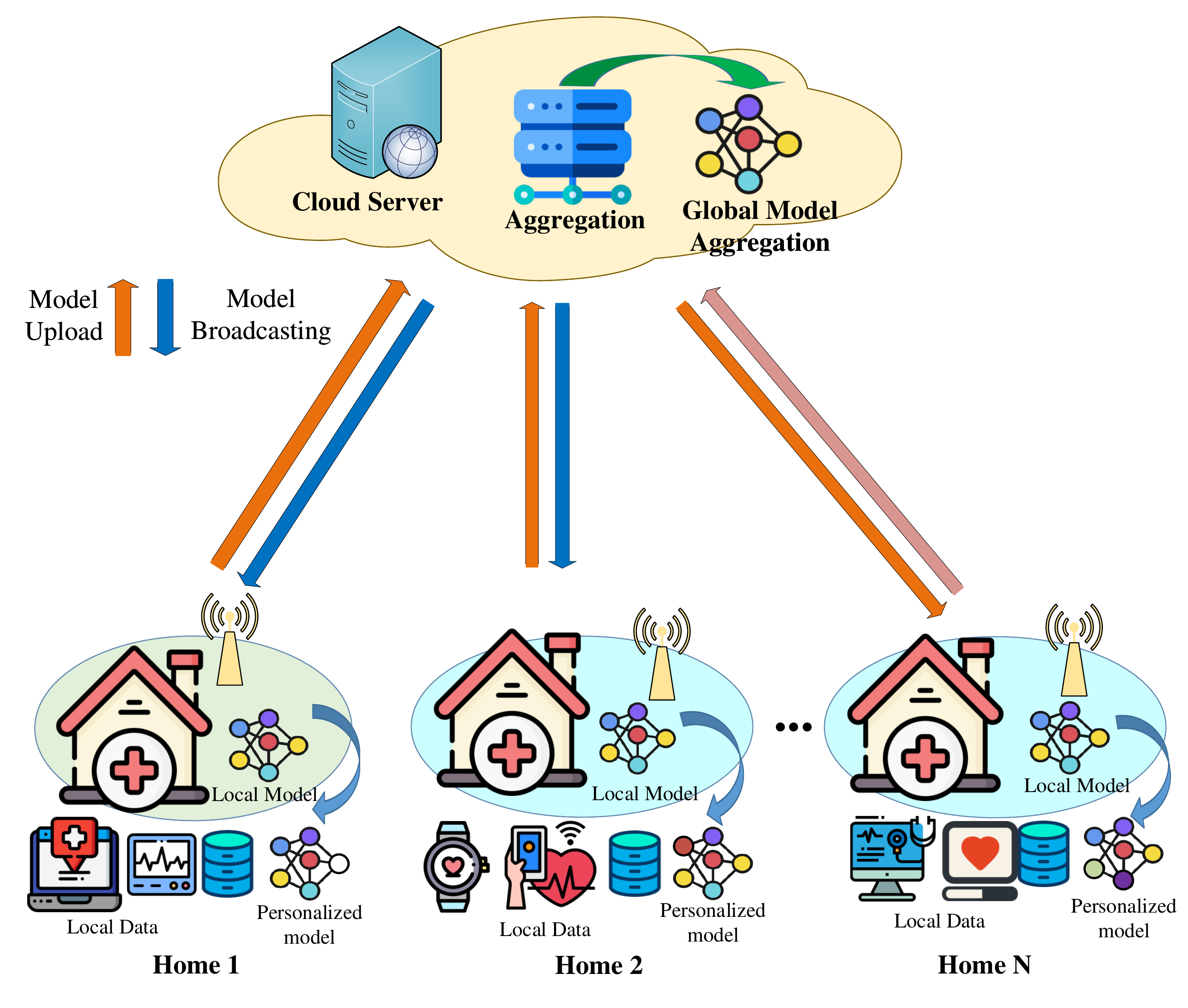}
	\caption{An edge-cloud architecture for personalized FL enabled in-home health monitoring services \cite{45}.}
	\label{Fig:FedHome}
	\vspace{-0.1in}
\end{figure*}

One of the very first works focusing on personalized FL for healthcare services is developed in \cite{45}. This work firstly points out critical challenges related to personalized FL for healthcare services. The first challenge is that different people naturally have different physical characteristics, the kids' body temperature is typically higher than that of adults and adults have very few falling steps when walking and doing exercise, thus fewer falling samples in the training dataset. Another challenge is that deploying a single global model may not be suitable for a specific person. To overcome these challenges, \cite{45} proposes an edge-cloud FL architecture, namely FedHome, for in-home healthcare services, where each local model is trained at home located at the network edge and the cloud is responsible for global model aggregation, as shown in Fig.~\ref{Fig:FedHome}. \textcolor{black}{The training at the cloud server mainly relies on the distributed datasets, which may be different among different homes. The shared model only captures the common features of all health users, and thus it may perform poorly on a particular user. As a result, to achieve personalized in-home health monitoring, each user integrates the trained global model with his personal health data.} To capture the personalized services, in-home IoMT devices download the global model from the cloud and trains its personalized model. The global model from the cloud is also used to generate a balanced dataset for all the devices in the network, thus decreasing the discrepancy between personalized models and the global model. Tested on the human activity recognition dataset, FedHome with the generative convolutional autoencoder at both the cloud and edges, can outperform several benchmarks in terms of training accuracy and model size. For example, FedHome can achieve the accuracy of 95.87\% while the personalized FL with \textcolor{black}{multi-layer perceptron} (MLP) and convolutional neural network (CNN) can only achieve  92.31\% and 91.77\%, respectively.  \textcolor{black}{The potential of this FedHome model can be extended to other IoT applications such as personalized human activity in hospital settings and personalized driving in smart transportation \cite{ yi2017personalized}. However, this personalized FL model also has limitations in terms of the imbalance in data distributions between the local data and the global data population. The reason for this is that users' personalized data is usually insufficient.}
The work in \cite{gudur2020resource} identifies a new challenge in FL design, which is called label heterogeneity. Concretely, each FL device has its own definition of data labels and learning model, which is independent of the definitions in other devices and the central server. 
However, such a new challenge typically exists in health and activity tracking scenarios. For example, the blood type is labeled as bloodtype in the device A; however, which is labeled as  bltype in the device B. This new challenge is solved in \cite{gudur2020resource} by devising an $\alpha$-weighted update. Particularly, the overlapping information of labels of different devices is aggregated at the central server. The proposed \textcolor{black}{approach} is examined on the \textcolor{black}{Animals-10 dataset \cite{dataset1}}, showing that the classification accuracy is increased by about 16.7\%. This approach is also tested on the human activity recognition dataset with an average increase of 9.153\% 11.01\% in \cite{gudur2020resourceconstrained} and \cite{krishna2020federated}, respectively. Recently, the work in \cite{rudovic2021personalized} develops a personalized FL framework to predict the pain by using face images. Each device employs a lightweight CNN model and keeps the learning parameters and updates of the last layer locally. This is to 1) personalize the local pain estimation model and 2) increase the protection of data privacy against potential adversaries. The proposed model is evaluated by using  the \textcolor{black}{UNBC-McMaster Shoulder Pain dataset \cite{lucey2011painful}}, showing that the proposed personalized FL with privacy preserving outperforms the conventional FL approach, which shares all the learning updates with the aggregation server, which in turn increases the probability of data breach. 

\subsection{Lessons Learned}
\label{Subsec:LessonsLearned}
\begin{table*}
	\centering
	\caption{Summary of recent advances in FL designs for smart healthcare. }
		\label{Table:FL_Advances}
		\begin{tabular}{|P{1.7cm}|P{0.5cm}|P{1.0cm}|P{1.2cm}|P{1.7cm}|P{3.5cm}|P{3.5cm}|}
			\hline
			\textbf{Theme}& 	
			\textbf{Ref.} &	
			\textbf{FL type} &
			\textbf{FL clients}& 	
			\textbf{FL aggregator} &
			\textbf{Key contributions}&
			\textbf{Limitations / Potential directions}
			\\
			\hline
			\multirow{9}{*}{\makecell{Resource \\Allocation}} & 
			\cite{yang2020scheduling} & HFL & IoMT devices & Cloud server &	Impact of scheduling policies on the convergence rate. & Only three base scheduling policies are considered.  
			\\ \cline{2-7} &
			\cite{nguyen2020toward} & HFL & Smart devices & Edge server &  A resource allocation for multiple FL services. &	The proposed algorithms are not evaluated on the health dataset.  
			\\ \cline{2-7} &
			\cite{xu2021bandwidth} & HFL & Smart devices & Edge server & Each device runs only one FL service and bandwidth allocation schemes for intra- and inter-service is proposed. & The proposed algorithms are not evaluated on the health dataset.
			\\ \hline
			
			\multirow{9}{*}{Secure FL} & 
			\cite{li2021blockchain} & HFL /VFL & Smart devices & Serverless & Bockchain for decentralized FL updates and the impact of lazy nodes are studied. & \textcolor{black}{Blockchain may increase the computing requirements of smart devices.} 
			\\ \cline{2-7} &
			\cite{bonawitz2016practical} & General FL & FL users & Aggregation server & A seminal approach of secure aggregation for FL. & The proposed method is examined on the non-health dataset.
			\\ \cline{2-7} &
			\cite{so2021turbo} & General FL & FL users & Aggregation server & An efficient secure aggregation method namely Turbo-Aggregate. &	Only numerical results are reported.
			\\ \hline
	\end{tabular}
\end{table*}

\begin{table*}
	\centering
	\caption{Summary of recent advances in FL designs for smart healthcare (continued). }
		\label{Table:FL_Advances1}
		\begin{tabular}{|P{2cm}|P{0.5cm}|P{1.0cm}|P{1.2cm}|P{1.4cm}|P{3.7cm}|P{3.5cm}|}
			\hline
			\textbf{Theme}& 	
			\textbf{Ref.} &	
			\textbf{FL type} &
			\textbf{FL clients}& 	
			\textbf{FL aggregator} &
			\textbf{Key contributions}&
			\textbf{Limitations / Potential directions}
			\\
			\hline
			\multirow{11}{*}{\makecell{Privacy- \\enhanced \\FL}} & 
			\cite{wu2021incentivizing} & HFL & IoMT devices & Cloud server & A privacy-preserving FL approach with multi-dimensional cost. & The propose approach is evaluated via numerical simulations. 
			\\ \cline{2-7} &
			\cite{malekzadeh2021dopamine} &	HFL & hospitals & Cloud server & A method is proposed to balance the differential privacy guarantee and deep learning accuracy. &	Advanced encryption is needed for privacy enhancement. 
			\\ \cline{2-7} &
			\cite{38} &	HFL & Smart phones & FL server & A study on the tradeoff between privacy protection and training accuracy. & The convergence of FL algorithms has not been verified. 
			\\ \cline{2-7} &
			\cite{kerkouche2021privacy} & HFL & Hospital & Central server & A privacy-preserving resource allocation, namely FL-SIGN-DP, is developed. & AI techniques such as DRL can be adopted at the server to exploit the historical samples.  
			\\ \hline
			
			\multirow{9}{*}{\makecell{Incentive- \\aware \\FL}} & 
			\cite{sarikaya2019motivating} & HFL & IoMT devices & FL server & A Stackelberg game is proposed to stimulate the participant of IoMT devices. &	Results are reported using numerical results only.
			\\ \cline{2-7} &
			\cite{tangincentive} & Cross-silo FL & Hospitals & Data centre & A non-cooperative game approach is used to devise an incentive mechanism for resource allocation. & The model is developed based on the assumption of IID data among hospitals. 
			\\ \cline{2-7} &
			\cite{zhao2021efficient} & HFL & Smart device & FL server & A DRL-based approach to evaluate the contribution of data owners. & A reward model can be developed to encourage the participant of more FL users. 
			\\ \hline
			
			\multirow{8}{*}{\makecell{Personalized \\FL}} & 
			\cite{45} & HFL & In-home sensing devices & Cloud server & A edge-cloud FL architecture is proposed for in-home healthcare services. &	The convergence of the proposed FL algorithm has not been reported.
			\\ \cline{2-7} &
			\cite{gudur2020resource} & HFL & IoMT devices & FL server & An $\alpha$-weighted update is proposed to mitigate the effect of label heterogeneity. & The convergence of FL algorithms has not been verified. 	
			\\ \cline{2-7} &
			\cite{rudovic2021personalized} & HFL & Smart devices & FL server & A personalized FL approach is developed to predict the pain via face images. & The challenge caused by data imbalance should be further investigated. 
			\\ \hline
	\end{tabular}
\end{table*}

In this section, we have reviewed the recent advances in FL designs for smart healthcare, stemming from five main themes: resource-aware FL, secure FL, privacy-enhanced FL, incentive-aware FL, and personalized FL. Some promising lessons learned are summarized in the following. 
\begin{itemize}
    \item \textbf{Training datasets}: Various designs have been investigated for general FL scenarios that may be applicable for smart healthcare services. However, it is observed that almost all the methods proposed in the literature are evaluated using non-healthcare datasets such as MNIST and Fashion MNIST. Evaluating FL designs using related healthcare datasets and from the healthcare perspective is necessary to better improve prospective FL-enabled healthcare services. 
    
    \item \textbf{Performance tradeoff}: Numerous performance metrics have been used in the literature to design FL schemes for healthcare services. However, there is a tradeoff between metrics such as privacy and learning accuracy as experimented in \cite{38}. This result suggests that the FL framework should be designed for specific healthcare applications. This also necessitates efficient FL schemes to take into account multiple learning objectives for healthcare applications. 
    
    \item \textbf{Communication efficiency}: Exchanging extremely large health data among health organizations and between the distributed organization and the central site is very resource-consuming and may not be economics-practical. Therefore, it is necessary to develop bandwidth-efficient and compressed FL methods for healthcare scenarios. For example, a sparse ternary compression framework is proposed in \cite{sattler2020robust} to compress both uplink and downlink transmissions between the aggregation server and end FL participants. Such a compression method can be extended to healthcare scenarios, especially when the size of health data (e.g., X-ray, diagnosing results, and EMR photos) and model updates is extremely large. 
    
    \item \textbf{Multi and personalized FL services}: Multi FL services are promising as more data will be generated at the network edge and each IoMT device can collect various types of data from various data sources. Personalized FL services will also play an important role in enhancing healthcare services as different patients have different health profiles and purposes. Joint healthcare services with general FL services are promising and should be further investigated in the future.
\end{itemize}

We summarize this section in the taxonomy Table~\ref{Table:FL_Advances} and Table~\ref{Table:FL_Advances1} along with the key technical aspects of each reference work to provide more insights into the recently advanced FL designs for smart healthcare. 


%% file: Applications.tex
\section{FL Applications in Healthcare}
\label{Sec:Applications}
In this section, we explore the emerging applications of FL in smart healthcare, namely federated EHRs management, federated remote health monitoring, federated medical imaging, and federated COVID-19 detection and diagnosis. 

\subsection{FL for Federated EHRs Management}
In the past few years, AI/DL technologies have been widely used in the healthcare sector to gain insights into health issues and disease progression by learning digital medical information extracted from EHRs for facilitating diagnosis and assessments as well as promoting medical research activities. One of the challenges faced in such traditional AI techniques is privacy leakage during data analytics. Indeed, compared to other domains, EHRs data in healthcare systems are highly sensitive and private. The removal of metadata such as patient information is insufficient to preserve the privacy of patients, especially in the complex healthcare settings where multiple parties such as hospitals and insurance companies have access to the common healthcare database as part of their employment requirements, including data analysis and processing. 

In fact, FL is able to provide much more reliable solutions for intelligent data analytics in EHRs management by exploiting AI functions for supporting healthcare services while adequately preserving user privacy based on the cooperation of multiple entities, e.g., patients and healthcare providers. For instance, a privacy-aware and resource-saving collaborative learning protocol based on FL is introduced in \cite{36} for an EHRs analytic management system with the cooperation of multiple hospital institutions and a cloud server. Here, each hospital runs an NN using its own EHRs with the help of cloud computing. To ensure privacy for model parameters in the FL process, a lightweight data perturbation method is considered to perturb the training-related data, which thus can defend model memorization attack in the learning. Although attackers can obtain perturbed information of EHRs, it is hard to obtain or recover the original data. Furthermore, an FL-based approach is proposed in \cite{37} to predict hospitalizations for patients diagnosed with heart diseases using their historic EHRs. Specifically, health data from an EHR system consisting of patients' smartphones and distributed hospitals are trained locally with respect to demographic information such as age, gender, and physical characteristics. Afterwards, the trained model parameters are aggregated by the cloud server for building a unified prediction model based on a global support vector machine (SVM) classifier. This aims to predict the future hospitalizations of patients due to heart diseases for hospital resource management (e.g., future estimation of treatment facilities) without disclosing the private dataset. 
	
\begin{figure}
	\centering
	\includegraphics[width=0.75\linewidth]{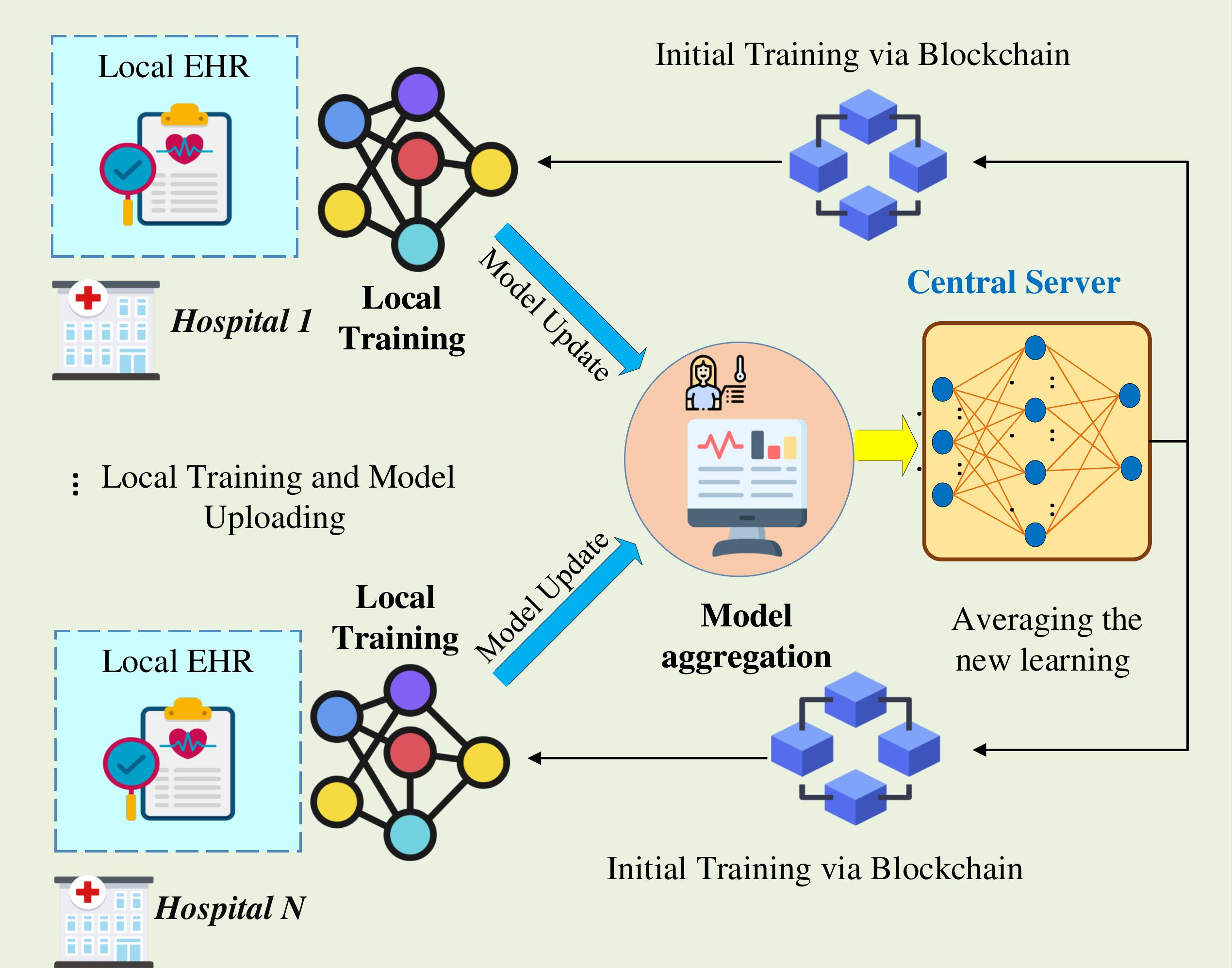}
	\caption{{Federated learning for collaborative EHRs analytics with blockchain. }}
	\label{Fig:FL_BC_EHRs}
	\vspace{-0.1in}
\end{figure}

To improve the privacy of FL-based healthcare, the work in \cite{38} employs FL to model the distributed EHR data learning among different hospital sites. A differential privacy-based solution  is adopted in \cite{506}, where each local hospital adds an artificial noise to mask the local updates before offloading them to the server for sensitive information protection. Three ML models are employed, including perceptron, support vector machine (SVM), and logistic regression, to perform FL training, showing that a competitive accuracy performance is achieved by using differential privacy-based FL methods compared to traditional FL schemes. A federated HER management solution is also considered in \cite{39}, where 58 different hospitals are employed to collaborate the prediction of patient mortality without moving health data out of their silos. A new FL-based method called SplitNN is studied in \cite{40} by implementing the vanilla FL configuration associated with split learning. \textcolor{black}{Conceptually, in split learning, a deep neural network is divided into multiple parts, each of which is trained on a different client. The trained data may reside at one client or at multiple clients, but no client can retrieve other clients' data. 
By splitting the network into multiple sections that are then transmitted to distributed clients, the network training can be implemented by offloading the weights of the split layer, while raw data sharing is not required.} 
For the case study in \cite{40},  each radiology client center trains a partial deep network and its neural parameters will be transmitted to a central server for model aggregation without sharing user information. To further improve the EHRs training performances in FL-smart healthcare, a statistic channel-based FL solution is considered in \cite{41}, where a small fraction of local gradients can be uploaded stochastically to the server from participating hospitals by choosing the channels with the most substantial variation. In other words, only the channel of neurons with a significant change in the training loop is used in the model aggregation, while ignoring neural channels with little change and less feature representation. A set of hospitals is adopted in simulations, comprising 30760 admissions with status information represented by alive or expired. A binary classification problem is formulated for mortality prediction using NNs in the FL simulation, showing an improvement in training convergence rates compared to traditional FL schemes. \textcolor{black}{Moreover, to support the federated mortality prediction, patient clustering is adopted in \cite{huang2019patient}, where patient subgroups capture similar diagnoses and geographical locations to train a neural network model to forecast mortality based on drug features. Another work in \cite{huang2020loadaboost} focuses on complexity reduction in FL settings for EHR mortality prediction, by applying an adaptive boosting method named LoAdaBoost for increasing the efficiency of federated machine learning in both IID and non-IID data distribution scenarios. } The study in \cite{42} considers EHRs training in the FL-based healthcare with recurrent NNs (RNNs) for predicting preterm-birth 3 months using structured datasets with uncertainty training. 42 hospitals are employed in the test, and each hospital performs the update of model generalizations as the local weights in the FL process. Other works in \cite{43}, \cite{44} also implement privacy-aware EHRs management solutions using FL for safe and reliable hospital networks in healthcare. A network of 20 hospitals is created in a case study in \cite{44} for building a federated EHRs management solution to improve the efficiency of health data analytics by the federated process achieved by the collaboration of medical centres, healthcare providers, and patients. Particularly, to provide security for data training, FL can be combined with blockchain  in the model update and data communications. As an FL example in Fig.~\ref{Fig:FL_BC_EHRs}, a number of hospitals cooperatively communicate each other via the blockchain to run DL algorithms by using local EHRs data. At each hospital, a DNN can be adopted for EHRs feature extraction and local modelling, where FL provides guidance for transmissions of model updates to perform the model aggregation at a common hospital. In this context, blockchain is used for secure model communication by providing an immutable data ledger managed by the hospitals and data centre \cite{nguyen2020blockchain}. The model update and transmission logs are recorded in the blockchain so that all entities can monitor the FL process in a transparent manner. 

\subsection{FL for Federated Remote Health Monitoring}
Recently, there is an increasing demand for developing smart solutions for administering remote healthcare from hospital-centric to home-centric. FL can be used for facilitating in-home health monitoring, by training a global model from distributed homes under the control of a server, e.g., a cloud server, while preventing data leakage by keeping user data locally \cite{45}. In this regard, the IoMT device (e.g., a user's mobile phone) at each home can learn a personalized model using convolutional neural networks (CNNs) by synthesizing a class-balanced dataset with its personal data and adjusting model gradients with the updated dataset in a fashion that the cloud and all homes update simultaneously. This not only addresses effectively imbalanced and non-IID data issues but also improves the personalized predictions. Extensive experiments are implemented using a realistic human activity dataset, showing that the FL-based approach can achieve a high accuracy of 95.41\%, an increase by 7.49\% over the standalone CNN scheme with low communication costs, in both the balanced and imbalanced data cases. 

Additionally, an FLT scheme is proposed in \cite{46} for wearable health monitoring where smart phones collaborate to train a shared CNN model with a cloud server for human activity recognition with privacy protection awareness. Since there is a large distribution divergence between models in the cloud and smartphones, transfer learning is utilized to make the training model more tailored which helps achieve personalization. Implementation results indicate the higher accuracy of the FL algorithm (5.3\% enhancement) for wearable activity recognition compared to traditional methods. The potential of this FL scheme can be expanded significantly in other interesting healthcare and medical applications, e.g., health monitoring, fall prediction, and disease diagnosis. Another remote healthcare monitoring system based on FL is designed in \cite{47} for obesity and comorbidity phenotyping control. To be clear, a two-stage federated natural language processing approach is proposed that allows the collaborative health data training using clinical notes from different hospitals or clinics without the need for sharing data. First, a patient representation model is built at each hospital for training a NN to predict the current procedural terminology code from the text of the notes. Then, a phenotyping ML model is built to perform collaborative training across multiple sites for the target phenotype for disease classification from three classes, namely presence, absence or questionable. Simulations for an FL setting with 10 hospital sites show a better precision and recall performance compared to non-FL approaches. 
	
Moreover, a mobile activity monitoring approach based on FL is considered in \cite{48} for supporting healthcare applications such as assisted living and fall detection. Here, a real-world heterogeneity human activity recognition dataset is employed which is distributed across mobile devices to perform federated NN training with four types of human activities, including sitting, walking, standing, and stairs up. A non-IID data environment is also established, where the activity data are split with disparities in both the predefined labels with these activity types and distributions in data. A disease prediction method is studied in \cite{49} by taking the advantage of FL using a large nationwide health insurance dataset distributed over 99 medical sites (e.g., hospitals and clinical labs) from 34 states in the US. The data include EHRs of diabetes, psychological disorders, and ischemic heart diseases. By comparing with conventional approaches such as centralized learning, local training without federation, the FL method achieves a competitive performance in terms of high accuracy rates and privacy protection. A new FL scheme called FedMood is suggested in \cite{50} for mood prediction and monitoring. The keyboard keystrokes such as the interval between two keystrokes are exploited for biometric identification that helps to predict depression through the analysis of keystroke habits of patients with depression. This is inherited from the fact that the depression patients often have different typing speeds compared to normal people. Particularly, mobile phones are employed to collect necessary information, e.g., key letters, special characters, and phone accelerometer values that can be trained via DNN and coordinated by a data server for  aggregation. A holistic experiment is conducted for both IID and non-IID data cases, indicating a good mood estimation accuracy (above 85\%) in comparison with local training and collaborative data sharing schemes. The work in \cite{51} also builds an FL-based health monitoring solution for analyzing the treatment effect of patients from the distributed hospital network. Interestingly, an entity called personalized treatment effect estimator is created in each hospital. Each estimator can be classified in each subgroup, where personal treatment effect includes the outcomes on patient characteristics and site indicator is used to estimate the global treatment effect at the coordinating site. 
\subsection{FL for Federated Medical Imaging}
Due to the privacy concerns, it is challenging to implement AI-based medical data imaging by fusing different medical institutions at a centralized entity. FL has emerged as a promising solution for supporting large-scale medical imaging tasks, by allowing to learn from multi-source datasets without the need for public data sharing. An FL approach is recently proposed in \cite{53} with a focus on solving the variations among clients (e.g., hospitals) by transforming the images of all clients onto a common image space via the federated process with a cloud-based global classifier. This enables to build a multi-source diffusion-coefficient image dataset for supporting automated classification. A generative adversarial network (GAN) is employed at each institution that can generate synthetic image datasets and translate its raw images to the target image space, which addresses cross-client variation problems with privacy preservation. Through the simulations on prostate cancer-related images, the proposed FL scheme can yield an accuracy score of 0.9722, achieving a performance enhancement by 0.13\% in comparison to non-FL schemes. 

Moreover, an FL model for federated brain imaging is also suggested in \cite{54}, aiming to support brain tumour segmentation using deep neural networks (DNNs). \textcolor{black}{Here, each federated client (e.g., MRI scan machine) has a fixed local dataset and reasonable computational resources to train the DNN structure and share the weight updates to the federated server for aggregation via a model averaging technique.} Although FL can protect user privacy leakage, it is still vulnerable to misuse risks such as training sample reconstructions at the server. Then, a differential privacy technique is adopted to add noise to each node's training process, distort the updates and mitigate the exposure of information during the model exchange. Simulations are conducted using a brain tumour dataset containing multiparametric pre-operative MRI scans of 285 subjects, indicating a similar segmentation performance with the ideal centralized scheme, while a degree of privacy protection is achieved. Another federated brain imaging approach is suggested in \cite{55} to take advantage of the magnetic resonance images (MRI) scans distributed across multiple clinical centres and institutions. In this case, an FL model is derived to simulate an end-to-end framework for data standardization, confounding factors correction, and measurement of variability of high-dimensional features, by the collaboration of medical sites and the central server. The MRI datasets along with covariates (e.g., age, sex information) are employed from 455 controls, 181 with non-progressive mild cognitive impairment (MCI), 208 progressive MCIc, 234 Alzheimer's disease, and 232 with Parkinson's disease. An approach using alternating direction methods of multipliers (ADMM) is also integrated for reducing the amount of iterations, aiming to mitigate training latency in the federated setting. In a similar direction, the authors in \cite{56} also consider a federated multi‑institutional collaborations for brain tissue-based MRIs analytics. A set of 10 medical institutions is involved where each entity runs an NN model to detect radiographically abnormal regions of each brain scanning image. 

To facilitate X-Ray scan imaging in smart healthcare, an FL-based methodology is proposed in \cite{58} for supporting diagnosis of acute neurological symptoms such as severe headache or loss of consciousness. Each hospital run a CNN-based DenseNet1212 model that can support feature propagation, encourage feature reuse, and minimize the number of neural parameters to train the X-ray image dataset provided by the Radiological Society of North America. To enhance privacy for FL-based medical imaging, differential privacy can be adopted by a Dopamine method \cite{59}. It is assumed that patients only trust their local hospital, and hospitals are non-malicious and non-colluding, while the server might be honest but curious. To this end, the hospitals need to provide a privacy levels to patients in each local update round in a fashion that the server can build a desired global model with high accuracy rates while data leakage is minimized. Each hospital adds a Gaussian noise of variance to calculate the privacy loss in a way that the effect of each update on the local momentum is the same as its effect on the aggregated model to strike the balance between the privacy cost and accuracy loss. A CNN-based SqueezeNet model \cite{60} is adopted to implement simulations, showing an increase of classification accuracy up to about 80\% on the real-world medical image dataset. A method called FL-based Magnetic Resonance Imaging Reconstruction (FL-MR) is proposed in \cite{61} for multi-institutional collaborations for MRI reconstruction. Here, the learned intermediate latent features from different hospitals are aligned with the distribution of the latent features at the target site. This is enabled by the collaborative training of local reconstruction networks at local sites and an adversarial domain identifier which aims to align the latent space distribution in the target domain. Moreover, a weakly supervised multiple instance learning approach is studied in \cite{62} based on the FL concept for gigapixel whole slide images in computational pathology. Here, in each hospital silo, the tissue regions are automatically segmented to extract the image patches, which are then embedded into a low-dimension feature representation using a CNN. Hen, the hospital performs training of a weakly-supervised learning model using its own whole slide image dataset with slide-level and patient-level labels as data features, before sending the trained gradients to the server for averaging. To demonstrate the feasibility of the proposed FL scheme, a weakly-supervised classification task is taken in histopathology for diagnosis of two separate disease datasets, i.e., renal cell carcinoma and breast invasive carcinoma. Simulations show that the FL approach can achieve the balanced accuracy of disease detection up to 90\% in various parameter settings, showing the potential of FL in decreasing the barriers to cross-institutional collaborations for facilitating pathology computation. \textcolor{black}{Another work in \cite{li2020multi} proposes an FL method for functional magnetic resonance imaging (fMRI) analysis, where a decentralized iterative optimization algorithm is designed with a randomization mechanism to coordinate the weight sharing process. A domain adaptation method is developed for cross-silo learning with labelled and unlabelled or semi-labelled images or text datasets. Simulation results show that the proposed model can boost neuroimage analysis performances and find reliable disease-related biomarkers by using multi-site data without data sharing. The real-world implementation of FL for medical imaging is also presented in \cite{roth2020federated} for breast density classification, showing that the FL scheme can perform 6.3\% on average better than standalone training approaches. }

\subsection{FL for Federated COVID-19 Detection and Diagnosis}
\begin{figure*}
	\centering
	\includegraphics[width=0.85\linewidth]{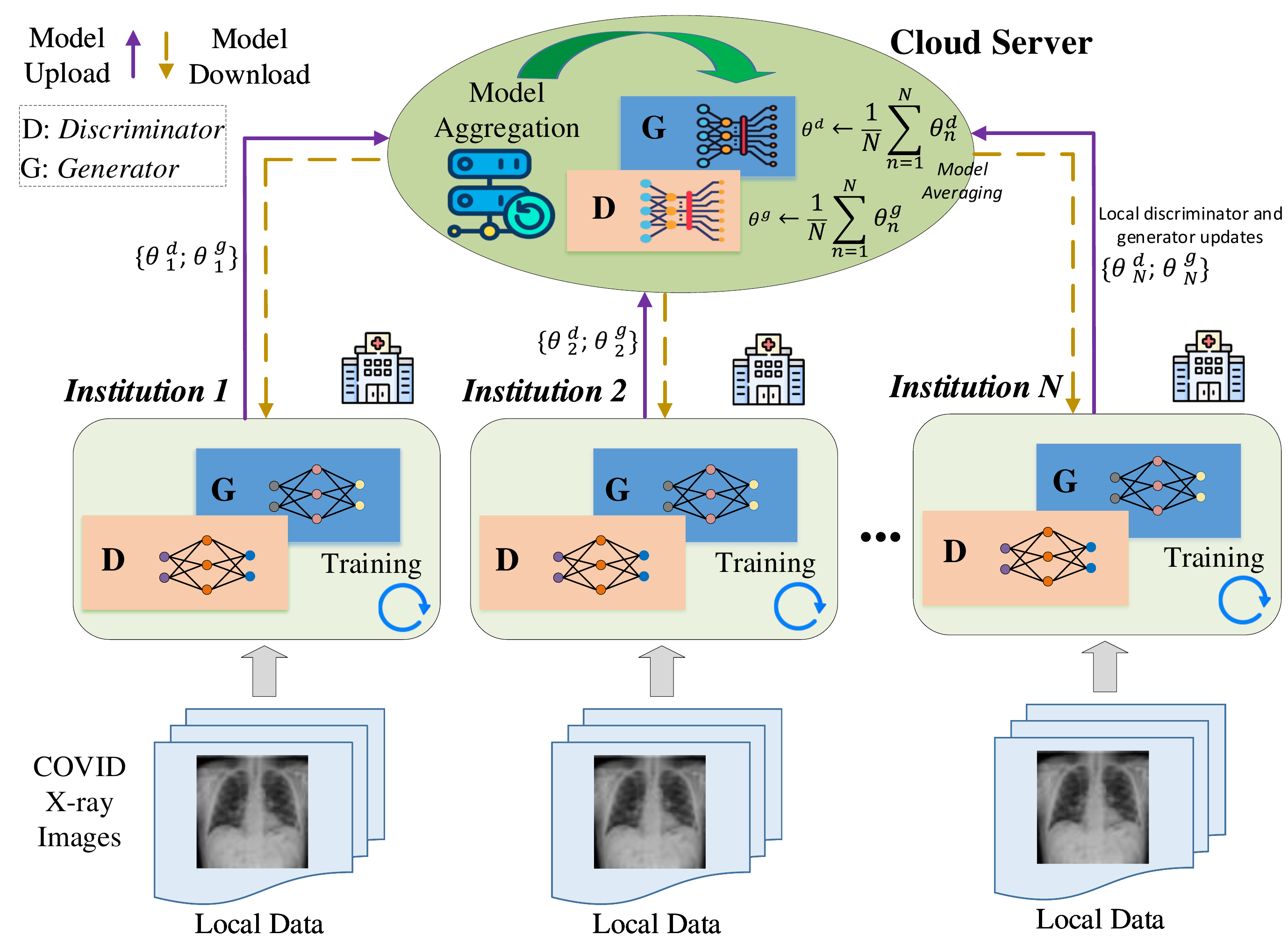}
	\caption{{An FL framework with GANs for COVID-19 detection.  }}
	\label{Fig:FL_COVID}
	\vspace{-0.1in}
\end{figure*}

Recently, COVID-19 has spread rapidly across the globe and become a major health concern of many countries \cite{63, pham2020artificial}. Many AI-based approaches have demonstrated significant promise for the early detection of COVID-19.  DL-based techniques such as convolutional neural networks (CNNs) \cite{pham2020artificial} have been widely used to identify COVID-19 cases by extracting essential features from chest X-rays. However, in the pandemic, collecting sufficient data to implement intelligent algorithms becomes more challenging and the user privacy concerns are growing due to the public data sharing with datacentres for COVID-19 data analytics \cite{67}. FL can be an ideal candidate for assisting COVID-19 detection, by its federation and privacy protection features. In this way, each institution participates in training their DL model using their local COVID-19 images, e.g., X-ray and computed tomography (CT) images, and only model parameters such as gradients are exchanged while there is no need to share actual data and sensitive user information. For example, a number of hospitals can cooperatively communicate via the blockchain to run CNN updates locally for identifying CT scans of COVID- 19 patients \cite{68}. At each hospital, a deep capsule network is developed to enhance image classification performance, while FL provides guidance for transmissions of model updates to perform the model aggregation at a data centre. Simulations from 34,006 CT scan slices (images) of 89 subjects verify a high COVID-19 X-ray image classification and low data loss in the FL algorithm running.

In a similar direction, FL is also used in \cite{69} to \textcolor{black}{ provide privacy-promoting AI solutions for COVID-19 chest X-ray image analytics.} Some preliminary experiments have been implemented, where multiple COVID-19 X-ray image owners run a CNN-based model for image classification, and then share the computed parameters with a data centre for mobile averaging while the data ownership of each user is guaranteed. Four state-of-the-art CNN models, e.g., MobileNet, ResNet18, MoblieNet and COVID-Net, are used in the federated setting for evaluation, where ResNet18 is proven with the best COVID-19 detection performance (98.06\%) in federated X-ray image learning settings. For instance, an FL framework for COVID-19 detection is illustrated in Fig~\ref{Fig:FL_COVID}. Each institution runs a local GAN consisting of a discriminator and a generator based on CNNs to learn the COVID-19 data distribution using its own local image dataset. Then, the local GANs synchronize and exchange the learned model parameters for aggregation at a cloud server, which then returns a new version of a global model to all institutions for the next training round. This process is repeated until a desired accuracy is achieved, aiming to generate realistic COVID-19 images for the detection of COVID-19.

Moreover, a dynamic fusion-based FL method is proposed in \cite{70} for CT scan image analysis to diagnose COVID-19 infections via two stages, including client participation and client selection. First, each client such as a medical institution make a decision to participate in the FL round based on the performance of the newly trained model. The central server also makes the decision to select which clients are permitted to update their local gradients by calculating the updating time. If a client cannot update its gradient within a predefined time interval, it is excluded from the FL aggregation. Another FL approach is designed in \cite{71} for COVID-19 screening from Chest X-ray images by the collaboration of multiple medical institutions. The focus is on Chest X-ray images classification to identify COVID-19 from non-COVID-19 cases, where the feature extraction and the classification of X-ray images are performed based on a CNN to detect the COVID-19 disease. At each hospital, a CNN model is trained by allowing each X-ray image to be put into a convolutional layer and output the probability of COVID-19 infection, and then a central server is used to aggregate synchronously with the local institutions for building a strong classification model for COVID-19 detection without compromising significantly user privacy which is valuable in the pandemic \cite{72}. Moreover, FL is also combined with DL to build a deep collaborative learning solution for detecting COVID-19 lung abnormalities in CT \cite{74}. The internal datasets are collected from a total of 75 patients confirmed COVID-19 infection at three local hospitals in Hong Kong for FL simulations, and then the generalizability is validated on external cohorts from Mainland China and Germany. 

\subsection{Lessons Learned}
The main lessons acquired from the review of the use of FL in smart healthcare applications are highlighted in the following.
\begin{itemize}
	\item  FL can be useful to build privacy-aware and resource-saving collaborative learning protocols for EHRs analytic management systems with the cooperation of multiple hospital institutions and a data centre \cite{36}. Interestingly, the use of FL opens new opportunities for federated EHRs analytics among distributed medical institutions despite the strict privacy regulation due to its learning nature by only allowing model parameters to be exchanged while raw data are kept at local sites. From \cite{38}, \cite{39}, \cite{42}, we can find that FL is a very useful learning approach to accelerate the accuracy rates of AI model training thanks to the use of distributed data resources and computation capabilities of multiple silos. 
	\item	FL is also useful for facilitating in-home health monitoring, by training a global model from distributed homes under the control of a data server, while preventing data leakage by keeping user data locally \cite{46}, \cite{47}. For example, FL can be employed to enable mobile activity monitoring \cite{48} for supporting healthcare applications such as assisted living and fall detection. Compared to conventional approaches such as centralized learning, local training without federation, we find that the FL method achieves a competitive performance in terms of high accuracy rates and privacy protection in smart health monitoring applications \cite{51}. 
	\item	Moreover, we also realize that FL plays a significant role in supporting medical imaging applications, by fusing different medical institutions at a centralized entity via the federated data training process \cite{53}. Particularly, FL is able to be deployed on medical devices such as MRI scan machine \cite{54} which has enough local dataset and reasonable computational resources to compute AI updates to join the FL process with the cloud server. Recently, the roles of FL in medical imaging is also investigated in X-Ray scan imaging \cite{58}, which has the great potential for supporting diagnosis of acute neurological symptoms such as severe headache or loss of consciousness. 
	\item	In the COVID-19 pandemic with growing privacy concerns, FL is particularly useful to support COVID-19 diagnosis and detection, by coordinating massive hospitals to build a common AI model \cite{68}. For example, we find that FL can be employed for COVID-19 screening from Chest X-ray images by the collaboration of multiple medical institutions. The feature extraction and the classification of X-ray images can be performed via collaborating among hospitals to detect the COVID-19 disease.	
\end{itemize}
In summary, we list the emerging applications of FL in smart healthcare in the taxonomy Table~\ref{Table:FL_Applications} and Table~\ref{Table:FL_Applications1} along with the key technical aspects of each reference work to provide more insights into the integrated FL-healthcare use cases.

\begin{table*}[h!t]
	\centering
	\caption{Taxonomy of FL applications for smart healthcare. }
	{\color{black}
		\label{Table:FL_Applications}
		\begin{tabular}{|P{1.7cm}|P{0.5cm}|P{0.6cm}|P{1.2cm}|P{1.4cm}|P{3.9cm}|P{4cm}|}
			\hline
			\textbf{Applied domain}& 	
			\textbf{Ref.} &	
			\textbf{FL type} &
			\textbf{FL clients}& 	
			\textbf{FL aggregator}&
			\textbf{Key contributions}&
			\textbf{Limitations}
			\\
			\hline
			\parbox[t]{20cm}{\multirow{15}{*}{\makecell{EHRs \\Management}}} & 
			\cite{35} & 	HFL  &	Hospitals &	Cloud server &	A collaborative learning with FL for EHRs processing. &	Convergence of the FL algorithm has 	not 	been verified.  
			\\ \cline{2-7}&
			\cite{36}& 	HFL  & Smart phones &	Data server&  	An FL scheme for predicting hospitalizations of patients.  &	The proposed model is simple and lacks detailed evaluations.  
			\\ \cline{2-7}&
			\cite{38} &	HFL &	Hospitals &	Data server&  	An differential privacy-based FL for federated EHRs training. &	The feasibility of differential privacy in real-world FL implementations should be considered. 
			\\ \cline{2-7}&
			\cite{40} &	HFL &	Radiology client center&	Cloud server&	An FL-based scheme with split learning for EHRs training.&	The proposed model is simple and training complexity is not verified. 
			\\ \cline{2-7}&
			\cite{42} &	HFL&	Hospitals&	Data server&  	Uncertainty FL for preterm-birth-related data analytics. &	The complexity of local training at each hospital should be analyzed. 						
			\\ \cline{2-7}
			\hline
			
			\parbox[t]{2cm}{\multirow{15}{*}{\makecell{Health \\Monitoring}}} & 
			\cite{45}&	VFL & Smart phones&	Cloud server&	A personalized FL scheme for remote activity recognition monitoring. &	Secure aggregation in FL communications has not considered. 
			\\ \cline{2-7}&
			\cite{46}&	FTL &	Wearable devices&	Cloud server&	A personalized FL scheme with transfer learning for human activity recognition. &	Communication costs and training complexity have not been verified. 	
			\\ \cline{2-7}&
			\cite{47} &	HFL&	Hospitals& 	Cloud server&	A two-stage federated natural language processing approach for obesity analytics. &	The privacy preservation for FL training should be considered. 
			\\ \cline{2-7}&
			\cite{48} &	VFL&	Mobile devices&	Cloud server&	An FL-based mobile activity monitoring approach for supporting healthcare applications. &	The comparison with other non-IID-aware FL techniques has been ignored. 
			\\ \cline{2-7}&
			\cite{49} &	HFL&	Medical sites&	Data centre&	An FL-based approach for large nationwide health insurance data analytics in the US. &	Issues related to federation agreement among medical sites need to be clarified in real-world settings. 
			\\ \cline{2-7}&
			\cite{50}&	HFL&	Mobile devices &	Data centre&	An FL-based method to predict mood using mobile devices. &	The training resource usage and privacy should be considered in mobile FL. 
			\\ \cline{2-7}
			\hline

	\end{tabular}}
\end{table*}

\begin{table*}[h!t]
	\centering
	\caption{Taxonomy of FL applications for smart healthcare (continued). }
	{\color{black}
		\label{Table:FL_Applications1}
		\begin{tabular}{|P{1.40cm}|P{0.5cm}|P{0.6cm}|P{1.2cm}|P{1.4cm}|P{4.1cm}|P{4.1cm}|}
			\hline
			\textbf{Applied domain}& 	
			\textbf{Ref.} &	
			\textbf{FL type} &
			\textbf{FL clients}& 	
			\textbf{FL aggregator}&
			\textbf{Key contributions}&
			\textbf{Limitations}
			\\
			\hline

			\parbox[t]{2cm}{\multirow{15}{*}{\makecell{Medical \\Imaging}}} & 
			\cite{53}&	VFL&	Hospitals&	Cloud server&	A variation-aware FL scheme for medical image construction. &	Learning accuracy has not been investigated. 
			\\ \cline{2-7}&
			\cite{54}&	HFL &MRI scan machines&	Federated server &	A privacy-enhanced FL scheme for brain imaging. &	FL convergence has not been investigated.  
			\\ \cline{2-7}&
			\cite{55}&	HFL&	Medical sites&	Data centre&	An FL-based MRI analytic framework for brain imaging. &	The practical aspects of the federated MRI training should be taken. 
			\\ \cline{2-7}&
			\cite{58}&	HFL&	Hospitals&	Data centre&	An FL scheme for supporting diagnosis of acute neurological symptoms. &	The proposed model is too simple and more simulations need to be done. 
			\\ \cline{2-7}&
			\cite{59} &	HFL&	Hospitals&	Data centre&	An FL-based medical imaging approach with differential privacy for collaborative secure training. &	Comparison of DL techniques in FL simulation has not been performed. 	
			\\ \cline{2-7}
			\hline
			
			\parbox[t]{2cm}{\multirow{15}{*}{\makecell{COVID-19 \\Detection}}} & 
			\cite{68}&	HFL &	Hospitals&	Data centre&	An FL scheme for collaborative COVID-19 detection. &	The proposed model is simple with a lack of detailed analysis.
			\\ \cline{2-7}&
			\cite{69}&	HFL &	Hospitals&	Data centre&	A number of experiments for COVID-19 detection with FL.&	The convergence of FL algorithms has not been verified. 	
			\\ \cline{2-7}&
			\cite{70} &	HFL&	Data clients&	Data centre&	A dynamic fusion-based FL method for CT scan image analysis to diagnose COVID-19 infections. &	Learning efficiency, e.g., latency, has not been verified.
			\\ \cline{2-7}&
			\cite{71} &	HFL&	Medical institutions &	Data centre&	An FL-based solution for COVID-19 screening from Chest X-ray images. &	Data loss caused by FL communications has not been considered.
			\\ \cline{2-7}&
			\cite{74}&	VFL&	Hospitals  &	Aggregator&	A federated deep learning method for detecting COVID-19 lung abnormalities in CT. &	Training latency has not been analyzed. 
			\\ \cline{2-7}&
			\cite{73} &	VFL&	Medical institutions &	Cloud server &	An FL method for COVID region segmentation in chest CT using multi-national data from China, Italy, Japan. &	Theorical analysis of FL communications and convergence should be provided. 				
			\\ \cline{2-7}
			\hline
						
	\end{tabular}}
\end{table*}

%% file: Projects.tex
\section{Real-world Projects of FL Implementation in Smart Healthcare}
\label{Sec:Projects}
In this section, we highlight several representative real-world projects of FL implementation in smart healthcare applications. 
\subsection{FL for Medical Imaging, US}
The feasibility of FL in medical imaging has been investigated via a real-world experiment conducted at the University of Pennsylvania and 19 other institutions worldwide in the collaborative healthcare project \cite{75}. Particularly, the Intel company has provided support to the FL-health project by leveraging the capabilities of Intel Xeon Scalable processors and Intel Software Guard Extensions (Intel SGX) for running FL functions at hospitals and the cloud server. This company also supports advanced DL algorithms along with strong hardware tools for accelerating the training, which helps building generalizable and state-of-the-art FL-healthcare models while improving the protection of sensitive patient data. \textcolor{black}{Several preliminary real-world experiments has been conduced at the centre of Biomedical Image Computing and Analytics of the University of Pennsylvania, showing that the FL-based approach can achieve an training accuracy rate of up to 90\% on image datasets, a competitive performance compared with the centralized scheme.} 
\subsection{FL for Healthcare Collaboration, UK}
As an effort to promote healthcare collaboration in the UK, the Nvidia corporation has recently joined the healthcare project with King's College London and Owkin to establish an FL platform for the National Health Service \cite{76}. The Owkin Connect platform running on NVIDIA Clara enables AI algorithms to be trained at local hospitals in the UK under the management of a central server. Particularly, blockchain has been integrated to provide traceability and monitoring of all healthcare data used for model training and data communications. The project is initially connecting four of London's premier teaching hospitals, providing federated AI services to support important medical domains such as cancer, heart failure and neurodegenerative disease. The project is expected to be deployed at least 12 hospitals around the UK in 2021. Also, the NVIDIA company is developing on-device AI platforms for deploying FL functions on smart devices such as wearables to handle medical image and video processing at high data rates in future large-scale federated healthcare projects. \textcolor{black}{Moreover, the UK Biobank, a large-scale biomedical study, has investigated the benefits of FL in real-world brain imaging under the support of the UK and French governments \cite{55}. The focus of this project is the use of FL in the data analytics for two diseases, including Alzheimer and Parkinson. The MRI datasets along with covariates (e.g., age, sex information) are employed from 455 controls, 181 with non-progressive mild cognitive impairment (MCI), 208 progressive MCIc, 234 Alzheimer’s disease, and 232 with Parkinson’s disease across multiple medical institutions in the UK. The preliminary results from the experiment shows that FL can improve the training performances in terms of better improved classification accuracy, which is highly applicable to medical image analytics in imaging-related disease diagnosis.}

\subsection{FL for International COVID-19 Project}
Recently, a real-world FL project for COVID-19 region segmentation in chest CT with the participation of medical institutions from China, Italy, and Japan is presented in \cite{73}. \textcolor{black}{Specifically, a multi-national database consisting of 1704 scans from these three countries is collected for setting up the FL framework, including 736 scans of 700 patients from the First Affiliated Hospital of Hubei University of Medicine in Hubei, China, 496 scans of 244 patients from the Self-Defense Forces Central Hospital, Tokyo, Japan, and 472 scans of 147 patients from San Paolo Hospital, Milan, Italy. The FL training performance is evaluated via the experiments of semi-supervised segmentation of COVID regions in 3D CT. Compared with the traditional local training approach using 945 scans with the support of expert radiologists, the FL-based approach is able to capture better the ground truth shapes and has less false positives with the collaboration of training resources from multiple medical centres. Given the strict regulatory policy on data privacy, FL is proven as a promising solution for countries to collaborate in the COVID-19 segmentation and detection without worrying the user information leakage and lacks of dataset.}

%% file: Challenges.tex
\section{Research Challenges and Future Directions}
\label{Sec:Challenges_Future-Directions}


This section presents the key challenges and future directions related to FL-Healthcare research. 


\subsection{Communication Issues in FL-based Smart Healthcare}
As an important part between FL users and the aggregation server, communication plays an important role in FL enabled healthcare services. Indeed, proper communication resource allocation schemes can significantly improve the learning performance. This becomes more important when a large number of IoMT devices need to connect with the aggregation server for model updates in the uplink and model broadcasting in the downlink.
In such a case, the aggregation server can employ efficient scheduling policies to select a suitable set of IoMT devices, as reported in many existing studies \cite{xu2021online, xia2020multi, luo2020hfel, yang2020scheduling}.
Another critical challenge from the communication perspective lies in the dynamic and fast variations of wireless channels, thus affecting the reliability and quality of learning updates between IoMT devices and the aggregation server. A possible solution is to take into account the effect of user dropping out \cite{bonawitz2016practical, so2021turbo} and considering more reliable design objectives such as outage probability and device availability. 

\subsection{Standard Specifications for Federated Healthcare Deployment}

Although many promising results of FL-enabled healthcare services have been shown in the literature, there is no standard and universal to evaluate the performance of different approaches for the same problem. For example, \textcolor{black}{various blockchain frameworks have been proposed for FL systems} such as removing the need for a central server and managing the reliability of local updates from IoMT devices. However, it is difficult to compare such approaches since they are proposed for different (healthcare) scenarios and different network settings/datasets are used to evaluate their performance. In addition, there are critical challenges related to the universal existence and standardization of communication protocols, device hardware, deployment scenarios, and aggregation methods. \textcolor{black}{Recently, a guideline on the architectural and design perspectives of FL is provided in IEEE Std 3652.1-2020 \cite{qiang2021white1}. Moreover, this guideline also presents major concerns about FL such as privacy, security, performance efficiency, and economic viability, as well as evaluation schemes and performance metrics of FL systems.}

\subsection{Quality of Federated Healthcare Training Data}
The training quality can be greatly degraded due to the heterogeneity of computational capabilities and data qualities of different hospital sites. A promising solution in this case is designing incentive mechanisms to motivate hospital/health organizations to use high-quality data for training as well as report reliable updates to the aggregation server. Game theory and blockchain are two important tools for designing incentive mechanisms \cite{xu2021bandwidth, kang2019incentive, 31, li2021blockchain, sarikaya2019motivating}. 
The training requirements (e.g., changes of data types, changes of learning rates, changes of training purpose-classification or regression) should be configured in a flexible manner so that the FL entity such as nurses, doctors, and patients can adjust their actions easily and appropriately. Such changes can affect the FL design and the underlying learning models (at the FL users and the aggregation server), which necessitates the development of adaptive FL approaches. AI is a promising tool as data from the past can be exploited to enhance the adaptability of FL model with future events. For example, DRL is leveraged in \cite{zhao2021efficient} to devise an incentive mechanism in case the conventional approaches perform worse when the feature dimensionality is sufficiently large. 

\subsection{Health Dataset Issues for Robust FL-based Health Data Analytics}
In realistic healthcare scenarios, different clients may have different datasets such as text, images, audio, and time series, and different data content such as blood type, heart rate, face images, and body temperature. Conventionally, almost all FL approaches in the literature are examined on a single dataset with a limited number of features. For example, the work in \cite{malekzadeh2021dopamine} is tested on the diabetic retinopathy dataset while the work in \cite{38} is evaluated on an EHR dataset, despite the fact that they are both proposed for privacy-preserving FL based healthcare services.
\textcolor{black}{Moreover, new heterogeneous FL approaches should be developed, where collaborating parties may have different models, but the central server can navigate this heterogeneity by private ensemble learning \cite{challenge4}.  In this work, an inference strategy is proposed to enable participants to operate an ensemble of heterogeneous models, without having to explicitly join the data at a single location.}

\subsection{FL-based Healthcare in Next-Generation Networks}
Although 5G networks are not fully available and commercially deployed over the world, there have been many research and development activities towards future 6G wireless systems \cite{chamitha2021frontiers6g}. 6G enables the availability of many applications such as Industry 5.0, intelligent healthcare, smart grid, holographic telepresence, and personalized body area networks. At the same time, many new technologies are introduced to meet much stricter 6G requirements such as blockchain, compressive sensing, THz and visible light communications, 3D networking, quantum communication, and large intelligence surface \cite{chamitha2021frontiers6g}. How to integrate FL functions on future 5G/6G medical devices, how to employ 6G devices, e.g., smart implants and wearables for large-scale FL-based healthcare, and what are new healthcare services enabled by 6G are open questions for future research. For example, future e-health services will be advanced by AI and FL capabilities, thus enhancing the quality of life and reducing the hospitalizations for patients \cite{mucchi20206g}.

{\color{black}
\subsection{FL with Provable Privacy Guarantee}
Despite the fact that FL has great potential in protecting the user data privacy, there are numerous privacy issues that should be addressed properly, especially in smart healthcare contexts due to the high sensitivity of the health-related data. According to \cite{11}, FL privacy issues can be categorized as membership inference attacks, unintentional information leakage, and generative adversarial network. For example, the attacker may misuse the global FL model to check the presence of a data sample in the FL health data set. Moreover, the patient's information can be inferred when the patient's device sends the local model updates to the central server located at the hospitals and health organizations. 
In order to design privacy-preserving solutions for FL-healthcare systems, making use of differential privacy, AI and advanced cryptography techniques is promising. The use of differential privacy to further enhance the privacy of FL systems is considered in many studies such as \cite{wu2021incentivizing, 38, malekzadeh2021dopamine, kerkouche2021privacy}. Such studies should be further investigated for healthcare applications such as how artificial noise is added to the model updates from the patient devices and to the global model from the central server. 
}

{\color{black}
\subsection{Security Issues in FL-based Smart Healthcare}
In FL-based smart healthcare systems, several participants at the client side can act as attackers and try to send poisonous model updates or fake information to degrade the model aggregation. Further, an adversary can contaminate information about data features during the local data training or modify local updates during the model transmission between local clients and the central server. At the server side, an external adversary can deploy attacks to steal information about the aggregated global model, which leads to serious privacy concerns such as information leakage. How to address such security issues is a real challenge in FL-based smart healthcare systems. Several solutions should be considered, such as using differential privacy \cite{challenge1} to protect training datasets against data breach. 
In addition, developing secure aggregation techniques \cite{challenge2} are promising solutions to provide a double-masking structure for encrypting local updates and implementing the key sharing among clients and the central server, aiming to protect clients against data modifications and attacks.

\subsection{Non-iidness and Data Quality in FL-based Smart Healthcare}
To achieve a desirable training performance in FL-based healthcare systems, a critical problem to be addressed is the non-iidness of the medical datasets which potentially makes the FL training divergent in the training. For example, a hospital can have a higher distribution for a certain type of local diseases than other hospitals in different geographic areas. In this case, the label distributions differ across medical institutions, making them challenging to join the federated data training. Without addressing this non-iidness issue, the data training would significantly suffer from quality degradation or even diverge. Solutions to overcome the non-iid challenge thus need to be developed, e.g., creating an additional subset of datasets to allocate fairly among clients \cite{35}, aiming to ensure efficient data training in FL-based smart healthcare. Another promising approach is to implement the feature shift among heterogeneous clients \cite{challenge3}, by using local batch normalization to adjust the feature distributions at the client side before averaging the local models. Quantitative metrics are needed to assess non-iid data in the FL-based smart healthcare sector, such as standard deviation, precision, and accuracy with respect to label/feature distribution skew and homogeneous partitions \cite{challenge5}.  }

%% file: Conclusion.tex
\section{Conclusions} 
\label{Sec:Conclusion}
FL is an emerging collaborative AI approach that has sparked an extreme interest to realize privacy-enhancing and scalable smart healthcare networks and applications. Given the lack of a comprehensive survey on the FL-healthcare topic in the open literature, this paper has provided a detailed survey of the use of FL in smart healthcare. We have first introduced the FL concept and motivations as well as the technical requirements for the utilization of FL in the smart healthcare domain. The recently advanced FL designs that would be useful to federated smart healthcare have been then discussed. Then, the key applications of FL in smart healthcare have been explored and discussed in details, including federated EHRs management, federated remote health monitoring, federated medical imaging, and federated COVID-19 detection. The emerging real-world projects related to FL-healthcare use cases have been provided, and the key lessons learned from the survey have been also highlighted. Finally, we have highlighted interesting challenges and discussed future directions in FL-smart healthcare. 

The application of FL in smart healthcare is still at its infancy and will quickly mature in the coming years for providing intelligent and privacy-enhanced health services. FL is expected to play a key role in realizing large-scale and collaborative healthcare systems and allow for a shift from centralized health data analytics to distributed healthcare operations with privacy awareness. We believe that this article will stimulate more attention in this emerging area and encourage more research efforts toward the full realization of FL in smart healthcare.